\documentclass[10pt,twocolumn,letterpaper]{article}

\usepackage{wacv}              

\usepackage{graphicx}
\usepackage{amsmath}
\usepackage{amssymb}
\usepackage{booktabs}
\usepackage{enumitem}
\usepackage{array,multirow,graphicx}
\usepackage{float}
\usepackage{pifont}
\newcommand{\cmark}{\ding{51}}%
\newcommand{\xmark}{\ding{55}}%
\usepackage{color, colortbl}
\definecolor{LightGray}{rgb}{0.92,0.92,0.92}
\definecolor{Red}{rgb}{1.0, 0.13, 0.32}
\definecolor{mygreen}{rgb}{0.0, 0.5, 0.0}

\usepackage{graphicx}
\usepackage{adjustbox}

\usepackage[pagebackref,breaklinks,colorlinks]{hyperref}

\usepackage[capitalize]{cleveref}
\crefname{section}{Sec.}{Secs.}
\Crefname{section}{Section}{Sections}
\Crefname{table}{Table}{Tables}
\crefname{table}{Tab.}{Tabs.}

\def\wacvPaperID{694} 
\def\confName{WACV}
\def\confYear{2024}

\begin{document}

\title{CLIPAG: Towards Generator-Free Text-to-Image Generation}

\author{Roy Ganz\\
Department of ECE\\
Technion, Haifa, Israel\\
{\tt\small ganz@campus.technion.ac.il}
\and
Michael Elad\\
Department of Computer Science\\
Technion, Haifa, Israel\\
{\tt\small elad@cs.technion.ac.il}
}
\maketitle

\begin{abstract}
Perceptually Aligned Gradients (PAG) refer to an intriguing property observed in robust image classification models, wherein their input gradients align with human perception and pose semantic meanings. While this phenomenon has gained significant research attention, it was solely studied in the context of unimodal vision-only architectures. 
In this work, we extend the study of PAG to Vision-Language architectures, which form the foundations for diverse image-text tasks and applications.
Through an adversarial robustification finetuning of CLIP, we demonstrate that robust Vision-Language models exhibit PAG in contrast to their vanilla counterparts.
This work reveals the merits of CLIP with PAG (CLIPAG) in several vision-language generative tasks. Notably, we show that seamlessly integrating CLIPAG in a ``plug-n-play'' manner leads to substantial improvements in vision-language generative applications. 
Furthermore, leveraging its PAG property, CLIPAG enables text-to-image generation without any generative model, which typically requires huge generators.
\end{abstract}

\vspace{-0.4cm}
\section{Introduction}
\label{sec:intro}

\begin{figure}[t]
    \centering
    \includegraphics[width=\linewidth]{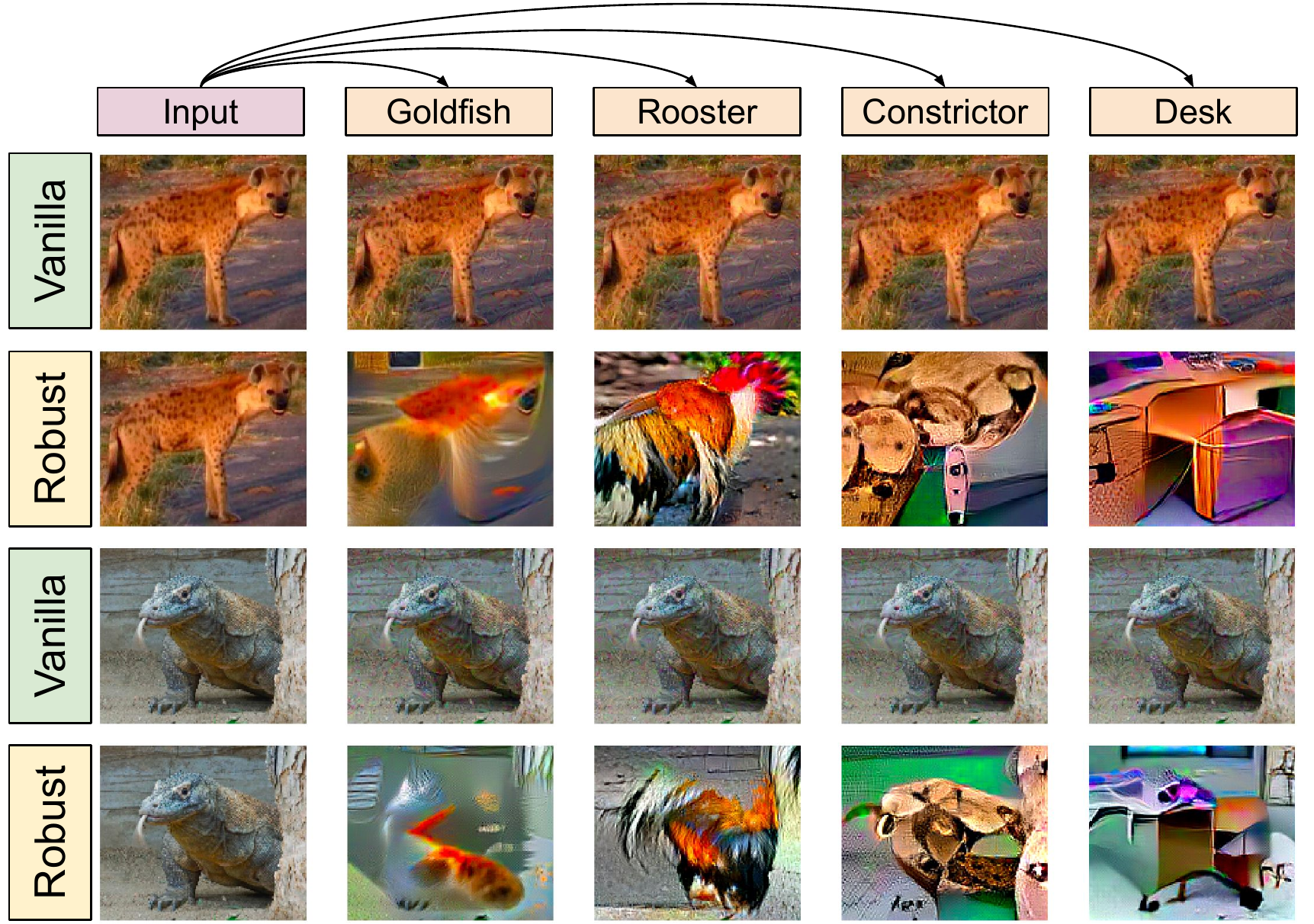}
    \caption{\textbf{Unimodal Perceptually Aligned Gradients}. Visualizations of large-$\epsilon$ targeted adversarial attacks on non-robust and robust ResNet-50, trained on ImageNet. Such attacks lead to semantically meaningful modifications in the robust model, indicating the generative capabilities of models with PAG. Contrary, the modifications done by the ``vanilla'' one are entirely meaningless.}
    \label{fig:pag}
    \vspace{-0.4cm}
\end{figure}

\begin{figure*}[t!]
    \centering
    \includegraphics[width=\textwidth,keepaspectratio]{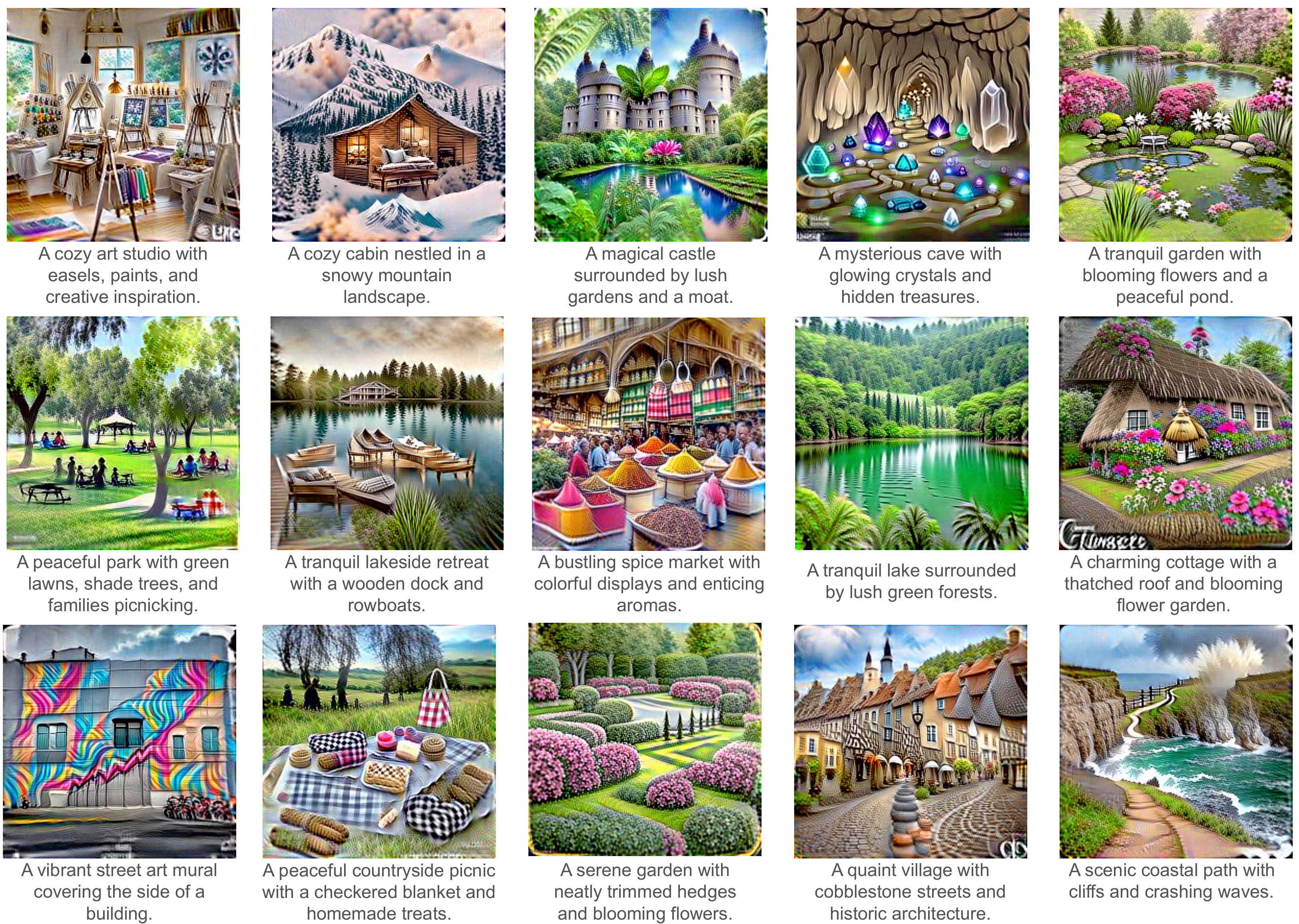}
    \caption{\textbf{CLIPAG generator-free text-to-image generation.}}
    \label{fig:generationExamples}
\end{figure*}

\textit{Adversarial robustness} is an essential objective in deep learning, requiring models to be insensitive to small malicious input perturbations, referred to as adversarial attacks. Tsipiras et al.~\cite{tsipras2019robustness} discovered a surprising property of adversarially robust models, commonly referred to as Perceptually Aligned Gradients (PAG). According to this trait, the input gradients of the model with respect to a specific class are semantically related to it, being significantly more aligned to human perception than non-robust ones.
An implication of this is that models with PAG have generative capabilities that can be leveraged using pixel space optimization.
Specifically, the outputs of strong targeted adversarial attacks lead to modifications that perceptually correlate with the target class (see Figure~\ref{fig:pag}). This exciting phenomenon has gained much research attention, with attempts at better understanding it~\cite{Engstrom2019AdversarialRA,general_property,DBLP:conf/icml/GanzKE23,srinivas2023models} and harnessing it for various computer vision applications~\cite{single_robust,ganz2022bigroc,DBLP:journals/corr/abs-2303-15409}. Interestingly, PAG has been explored so far solely in the context of unimodal vision-only applications.

\begin{figure*}[t!]
    \centering
    \includegraphics[width=\textwidth]{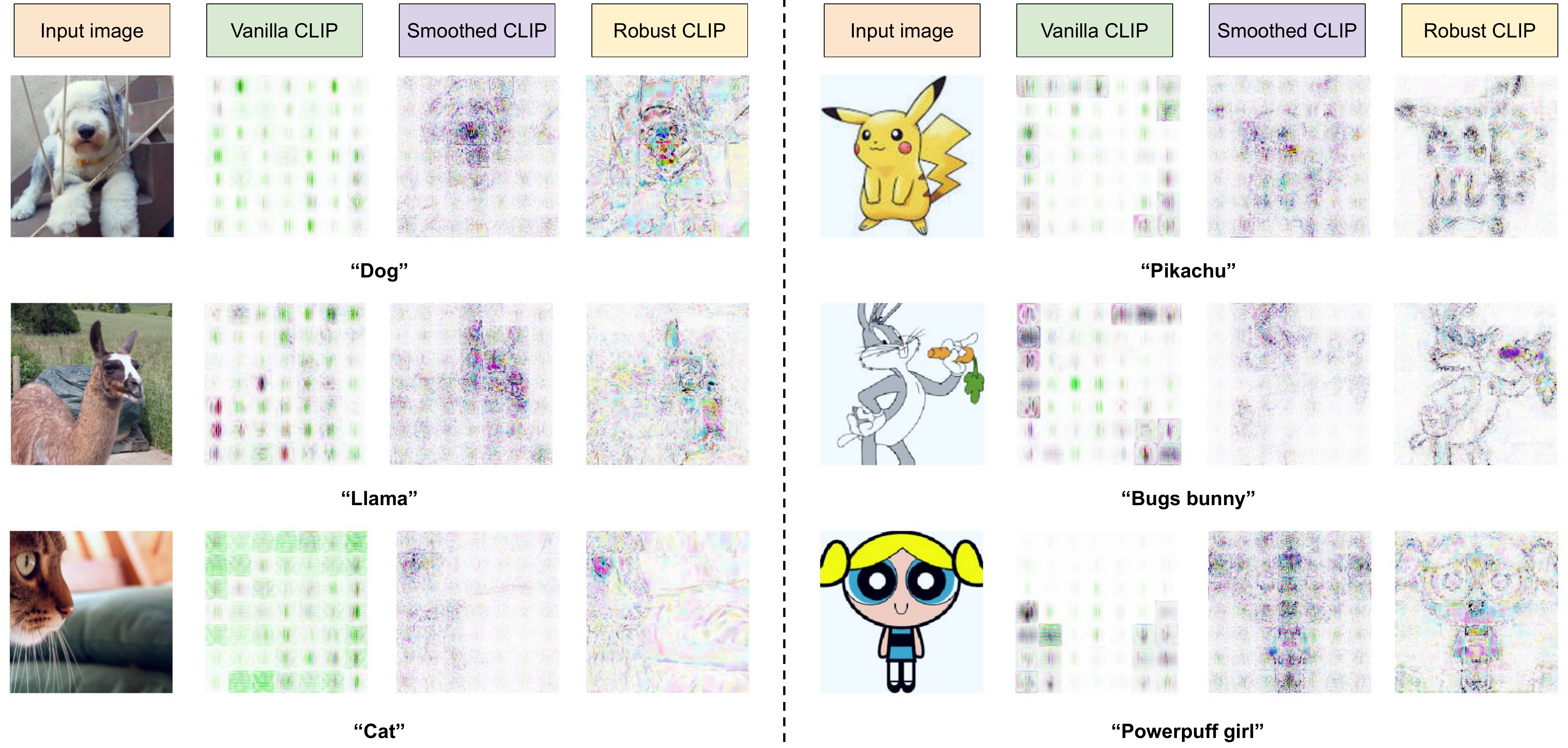}
    \caption{\textbf{PAG in CLIP.} A demonstration of the PAG phenomenon using CLIP. We present the input-gradients w.r.t. a given text of both the ``vanilla'' CLIP, randomized-smoothed CLIP, and CLIPAG in a zero-shot setting for natural images from ImageNet~\cite{deng2009imagenet} (left) and cartoons (right). While the former's gradients are completely meaningless, the latters lead to gradients that better align perceptually with the given text. Specifically, the adversarially robust CLIP leads to better alignment.}
    \vspace{-0.2cm}
    \label{fig:VL_PAG}
\end{figure*}

In contrast to the existing PAG literature that primarily focuses on unimodal vision-only applications, our work delves into the domain of Vision-Language (VL) models, which is gaining significant research and attention 
these days~\cite{clip,DBLP:conf/icml/0001LXH22,DBLP:journals/corr/abs-2301-12597,DBLP:conf/icml/WangYMLBLMZZY22,DBLP:conf/cvpr/ZhangLHY0WCG21,DBLP:journals/corr/abs-2305-17718,wang2022git,
DBLP:journals/corr/abs-2301-07389}.
Our exploration focuses on CLIP~\cite{clip} -- Contrastive Language-Image Pretraining -- a powerful Vision-Language model that learns a joint feature space for images and their captions. 
Building upon the understandings from the unimodal PAG research~\cite{cohen2019certified,general_property,tsipras2019robustness}, we consider 
an adversarial finetuning of the visual part of CLIP as a method that can potentially induce gradient alignment.
We demonstrate that while ``vanilla'' CLIP does not possess PAG at all, its robust counterpart does (see Figure~\ref{fig:VL_PAG}). 
We denote the resulting model as \textbf{CLIPAG} -- \textbf{CLIP} with \textbf{P}erceptually \textbf{A}ligned \textbf{G}radients, and show experimentally that adversarial training in this VL model implies PAG, as in unimodal ones.

Ever since its introduction, CLIP has become a foundational model for a wide range of text-to-image generative tasks~\cite{clipasso,clipdraw,clipstyler,styleclip,stylegan-nada,vqgan-clip,dalle2,glide,gigagan}. These often involve modifying images to maximize the alignment with a given text prompt, achieved by deriving CLIP's vision encoder and updating the image to maximize the cosine similarity with the text.
However, CLIP is known to be vulnerable to adversarial attacks~\cite{Fort2021CLIPadversarial} and lacks PAG, which poses a significant challenge in achieving the desired meaningful visual modifications. Indeed, CLIPDraw~\cite{clipdraw}, a CLIP-based text-to-drawing framework, acknowledged this limitation, stating ``\emph{synthesis through-optimization methods often result in adversarial images that fulfill the numerical objective but are unrecognizable to humans}''.
To mitigate this, researchers have developed ad-hoc techniques and tricks to regularize and improve CLIP gradients, such as optimizing a generator's latent space~\cite{styleclip,stylegan-nada,clipstyler,vqgan-clip} or utilizing Bézier curves rather than operating in the pixel-domain~\cite{clipdraw,clipasso}. In this context, our proposed CLIPAG is a natural solution to this limitation, as discussed next.

We embark on demonstrating the benefits of CLIPAG by seamlessly integrating it into existing CLIP-based generative frameworks in a ``plug-n-play'' manner. Specifically, we consider both text-to-image generative tasks using CLIPDraw~\cite{clipdraw} and VQGAN+CLIP~\cite{vqgan-clip} and text-based stylization using CLIPStyler~\cite{clipstyler}.
We show that replacing CLIP with CLIPAG leads to improved performance in all these fronts. 
Interestingly, CLIPAG alleviates the need for gradient regularization techniques, offering a more straightforward approach for leveraging CLIP in such tasks.

Inspired by the above, we propose a novel text-to-image generation via a simple iterative framework using CLIPAG. Unlike the above-described experiments in which CLIPAG is merged into existing solutions, this synthesis framework is a direct \emph{pixel-domain-based} approach. 
Amazingly, and in contrast to existing text-to-image methods that rely on huge generative networks (\textit{e.g.}, diffusion and GANs), CLIPAG enables high-quality pixel-space image generation (see examples in Figure~\ref{fig:generationExamples}) without any explicit training of a generative model and while using a small pretrained network. 


To summarize, the main contributions of this work are as follows: (i) 
We introduce CLIPAG - an adversarially finetuned version of the visual encoder of CLIP that exhibits perceptually aligned gradients; 
(ii) We integrate CLIPAG into existing text-to-image frameworks, substantially improving their performance while also simplifying these algorithms; and  
(iii) We leverage CLIPAG to propose a simple, generator-free text-to-image synthesis solution, producing high-quality synthesized images.

\section{Related Work}
\label{sec:related}
\subsection{Adversarial Attacks and Robustness}
\paragraph{Adversarial Attacks}
Given an image classifier $f_\theta(\mathbf{x})$, adversarial examples are crafted by attackers in order to fool it and divert the classification decision.
It was discovered that adding a small imperceptible noise $\delta$ to an image can lead to misclassification~\cite{szegedy2014intriguing,goodfellow2015explaining}. 
In practice, the adversarial noise is constrained within a defined threat model $\Delta$, often defined as a small norm ball (\textit{e.g.},
$\Delta = \{\delta \ : \ \lVert\delta\rVert_\infty \leq \frac{8}{255}\}$).
Formally, given an input sample $\mathbf{x}$, its true label $y$, and a threat model $\Delta$, a valid adversarial example $\hat{\mathbf{x}}$ satisfies the following conditions: $\hat{\mathbf{x}} = \mathbf{x} + \delta \ s.t. \ \delta \in \Delta$ and $y_{\text{pred}} \neq y$, where $y_{\text{pred}}$ is the predicted label by the classifier $f_\theta$ for $\hat{\mathbf{x}}$. The process of generating such examples is called adversarial attacks, and numerous methods have been developed for this purpose~\cite{goodfellow2015explaining,carlini_wagner,boosting_attacks,madry_pgd}. This work focuses on the Projected Gradient Descent (PGD) method~\cite{madry_pgd}.
\vspace{-0.5cm}
\paragraph{Adversarial Robustness}
The above-described vulnerability of classifiers 
sparked research dedicated to enhancing their robustness 
against such attacks. 
A commonly considered solution is adversarial training~\cite{goodfellow2015explaining,madry_pgd}, which approximates the solution of the following min-max optimization:
\begin{equation}
    \label{eq:adv_train}
    \min_{\theta} \sum_{(\mathbf{x},y)\in \mathcal{D}} \max_{\delta \in \Delta} \mathcal{L} (f_{\theta}(\mathbf{x}+\delta),y),
\end{equation}
in which the classifier is trained to correctly classify the most challenging adversarial examples allowed by the threat model $\Delta$.
An additional effective technique for robustifying neural networks is randomized smoothing~\cite{cohen2019certified}, in which the classifier is smoothed by convolution with Gaussian noise. Specifically, in the $L_2$ case, the robust classifier $\hat{f}_{\theta,\sigma}$ is a smoothed version of $f_\theta$,
\begin{equation}
    \label{eq:randomized_smoothing}
    \hat{f}_{\theta,\sigma} = \mathbb{E}_{\mathbf{n}\sim\mathcal{N}(0,\sigma^2\mathbf{I})}[f_\theta(\mathbf{x}+\mathbf{n})],
\end{equation}
where $\sigma$ controls the robustness-accuracy tradeoff.

\subsection{Perceptually Aligned Gradients}
Perceptually aligned gradients (PAG) \cite{tsipras2019robustness, Engstrom2019AdversarialRA, etmann19a} refer to classifier input-gradients, $\nabla_{\mathbf{x}} f_\theta(y|\mathbf{x})$, that are semantically aligned with human perception. 
Consequently, when an image is altered to maximize the probability of a specific class in a model with PAG, the modifications made to the image are semantically meaningful, as demonstrated in Figure~\ref{fig:pag}.
PAG has been found to exist in adversarially trained models but not in ``vanilla'' ones \cite{tsipras2019robustness}, indicating that the features learned by the former are more aligned with human vision. 

PAG has recently drawn significant research attention. 
Theoretically oriented studies have focused on better understanding this trait and the circumstances for its appearance. By demonstrating PAG in randomized smoothed classifiers, it has been established that PAG is a general property of robust models and not solely an artifact of adversarial training~\cite{general_property}. Furthermore, Ganz \textit{et al.}~\cite{DBLP:conf/icml/GanzKE23} demonstrated the bidirectional relationship between PAG and robustness, revealing that PAG implies robustness and vice-versa. Additionally, Srinivas \textit{et al.}~\cite{srinivas2023models} recently shed some light on the cause of PAG via off-manifold robustness analysis.
Applicative-oriented studies aimed at leveraging PAG for generative tasks, such as image generation and image-to-image translation~\cite{single_robust}, improving state-of-the-art image generation results \cite{ganz2022bigroc}. PAG has also been explored for improved robust classification~\cite{DBLP:journals/corr/abs-2303-15409}.

Despite this extensive line of work and the growing interest in multimodal networks among the computer vision research community, PAG has only been studied within the context of unimodal vision-only architectures and has never been explored in the vision-language domain. 
In this work, we aim to close this gap and investigate the existence of PAG in Vision-Language multimodal architectures and its connection to adversarial robustness using both adversarial training and randomized smoothing.
Armed with this, we explore the potential of multimodal PAG in improving multiple text-to-image generative applications.

\begin{figure}[t]
    \centering
    \includegraphics[width=\linewidth]{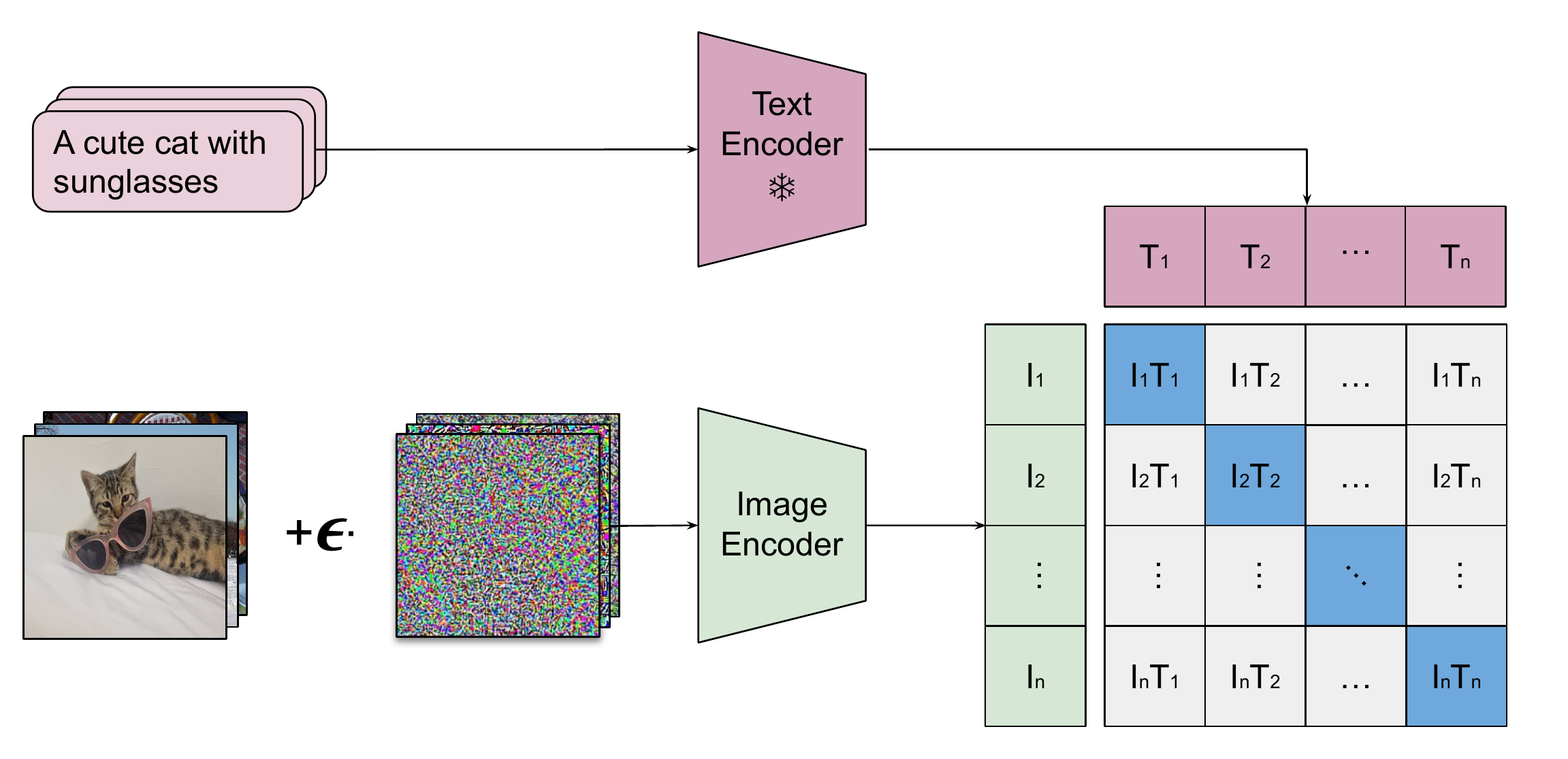}
    \caption{\textbf{Vision-Language Contrastive Adversarial Training.} According to Equation \ref{eq:vl_adv_train}, we first craft adversarial perturbation $\delta$ to minimize the similarity between the corresponding image-text pairs, \textit{i.e.}, reduce the values on the main diagonal of the depicted matrix. Next, we train the CLIP model to align the adversarial examples and their matching text, \textit{i.e.}, to maximize the values on the main diagonal. Repeating these steps results in adversarially robust CLIP that possesses Perceptually Aligned Gradients.}
    \label{fig:method}
    \vspace{-0.5cm}
\end{figure}

\subsection{CLIP in Vision-Language Generative Tasks}
CLIP (Contrastive Language-Image Pretraining)~\cite{clip} is a multimodal Vision-Language model pretrained to align a massive corpus of $400$ million pairs of images and their captions. 
The outstanding richness of CLIP's visual and textual space has been leveraged for various text-to-image generative tasks.
Several studies have focused on image generation conditioned on textual descriptions, mainly by guiding the visual result to be aligned with the given text in CLIP's space~\cite{clipdraw,vqgan-clip,gigagan,glide,dalle2,latentdiffusion}.
VQGAN+CLIP~\cite{vqgan-clip} proposed a training-free method to generate images from text by optimizing the latent code of a pretrained VQGAN~\cite{yu2021vector} to output an image that matches the textual description in the CLIP space.
Similarly, Ponce \textit{et al.} introduced ClipDRAW~\cite{clipdraw}, a method that combined CLIP and Bézier curves for generating drawings from texts by optimizing the parameters of such curves for best alignment.
Interestingly, \cite{clipdraw} has identified CLIP's susceptibility to adversarial attacks: ``\emph{A key issue in synthesis through optimization is that the produced images often leave the space of natural images, or fool the system through adversarial means}''. Thus, such works mitigated this by avoiding pixel-domain optimization and performing multiview augmentations.

Another line of work involving CLIP is text-guided style transfer and image editing.
StyleCLIP~\cite{styleclip} and StyleGAN-NADA~\cite{stylegan-nada} proposed to leverage a pretrained StyleGAN~\cite{karras2020analyzing} model with CLIP to adjust the style of images to match a given textual descriptions.
CLIPStyler~\cite{clipstyler} tackles a similar task using a framework that consists of several CLIP-based losses, augmentation pipes, and a style network being trained for each image.
CLIP was also used for localized text-based image-editing using internal learning~\cite{bartal2022text2live}.
These studies demonstrate the broad applicability of CLIP in text-to-image generative tasks. 

\begin{figure*}[t]
    \centering
    \includegraphics[width=0.95\textwidth]{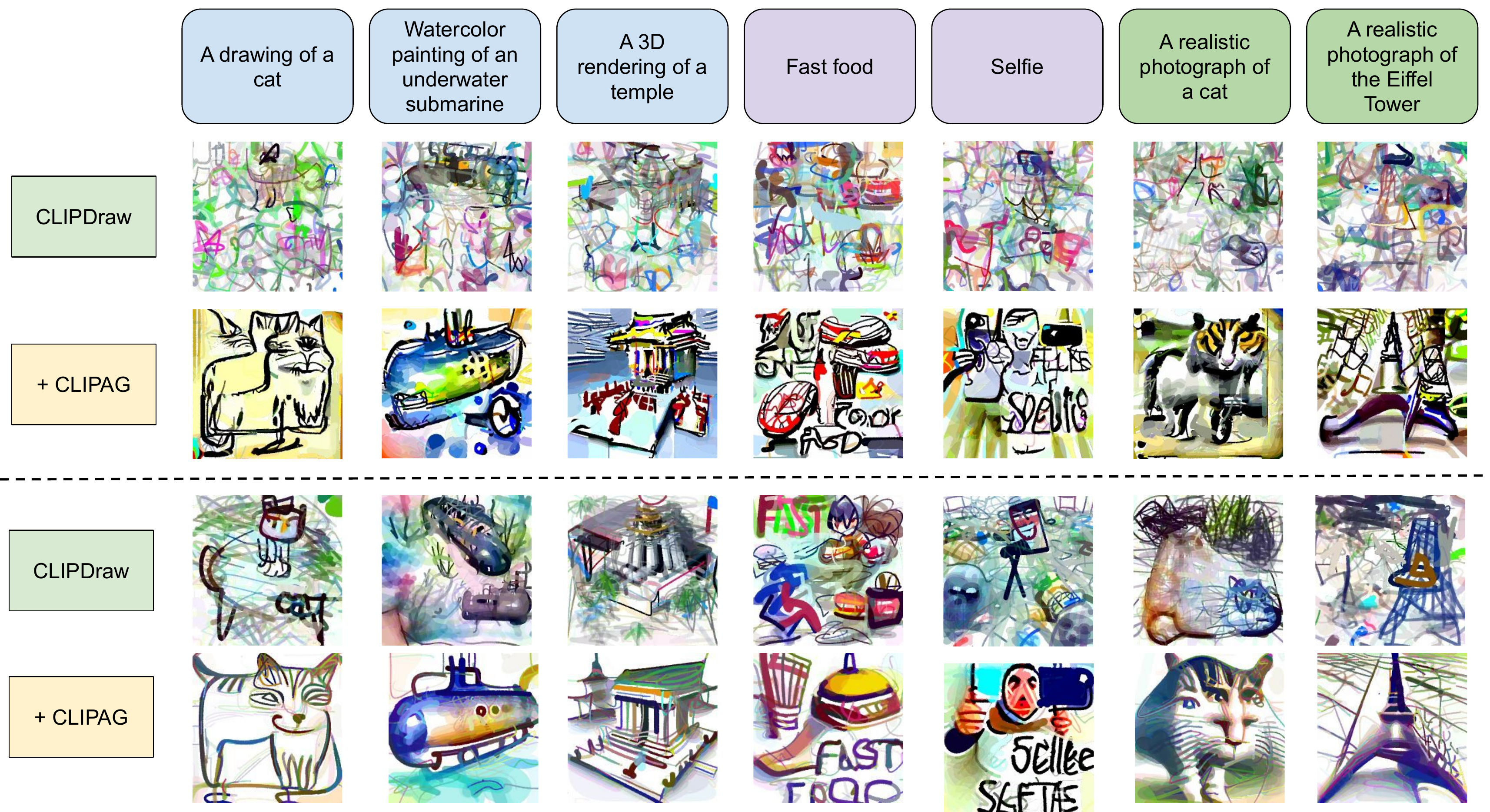}
    \caption{\textbf{CLIPDraw results.} Visualization of CLIPDraw outputs with CLIP and CLIPAG using different textual prompts styles (artistic in blue, abstract concepts in purple, and realistic in green. The top two rows represent results without augmentation. As can be seen, in this case, CLIP completely fails to guide the optimization process toward meaningful outputs. However, CLIPAG significantly outperforms it, leading to improved drawings that align with the textual description. Moreover, when applying augmentations (bottom rows), CLIPAG still leads to better visual outputs, as also indicated in the quantitative evaluation presented in Table~\ref{tab:clipdraw}.}
    \label{fig:clipdraw}
    \vspace{-0.5cm}
\end{figure*}

\section{Obtaining Vision-Language PAG}
\label{sec:method}

In this paper, we delve into the concept of Perceptually Aligned Gradients (PAG) within vision-language models. We focus on the image encoder's input gradients with respect to a given textual input, aiming for structured content that is semantically correlated with the text. 
To this end, we leverage the well-established observation that both adversarial training and randomized smoothing lead to aligned gradients in unimodal vision-only models~\cite{tsipras2019robustness,general_property}. 
First, we explore the gradients of ``vanilla'' CLIP, using the publicly available CLIP ViT-B/32~\cite{ilharco_gabriel_2021_5143773}, using both natural images from ImageNet~\cite{deng2009imagenet} and arbitrary cartoon ones.
To this end, we obtain input gradients by deriving the image encoder to maximize the cosine similarity with a given text in CLIP's feature space.
As can be seen in Figure~\ref{fig:VL_PAG}, the standard CLIP model has no alignment with the semantically meaningful features while also exhibiting a strong blockiness effect due to the Vision Transformer (ViT) architecture~\cite{dosovitskiy2021image}.

Next, we examine the gradient alignment of the same CLIP model with randomized smoothing.
This approach mitigates the blockiness effect and improves the alignment, as seen in Figure~\ref{fig:VL_PAG}, but only to some extent.
Hence, we propose to adversarially finetune the CLIP model to improve the PAG property. To this end, we 
adopt adversarial training techniques~\cite{madry_pgd}, as illustrated in Figure~\ref{fig:method}. We denote the Text Encoder and Image Encoder as $f^T_{\theta_T}$ and $f^I_{\theta_I}$, respectively, where $\theta_T$ and $\theta_I$ represent the models' parameters.
Given an image $\textbf{x}$ and its corresponding caption $t$ from an image-text dataset $\mathcal{D}$, we first craft adversarial example $\mathbf{x + \delta}$ aimed to minimize the similarity between matching image-caption pairs. Subsequently, we update the model weights to maximize the similarity between the adversarial examples and their corresponding captions.
Formally, we propose solving the following optimization problem, which extends adversarial training to the multimodal case: 
\begin{equation}
    \label{eq:vl_adv_train}
    \min_{\theta_I,\theta_T} \sum_{(\mathbf{x},t)\in \mathcal{D}} \max_{\delta \in \Delta} \mathcal{L}_{SIM} (f^I_{\theta_I}(\mathbf{x}+\delta),f^T_{\theta_T}(t)),
\end{equation}
where $\mathcal{L}_{SIM}$ represents cosine similarity loss calculated in CLIP's feature space. We visualize the optimization process described in Equation~\ref{eq:vl_adv_train} in Figure~\ref{fig:method} while taking into account the fact that CLIP is trained over batches of pairs via 
 contrastive learning.

\begin{figure*}[t]
    \centering
    \includegraphics[width=0.87\linewidth]{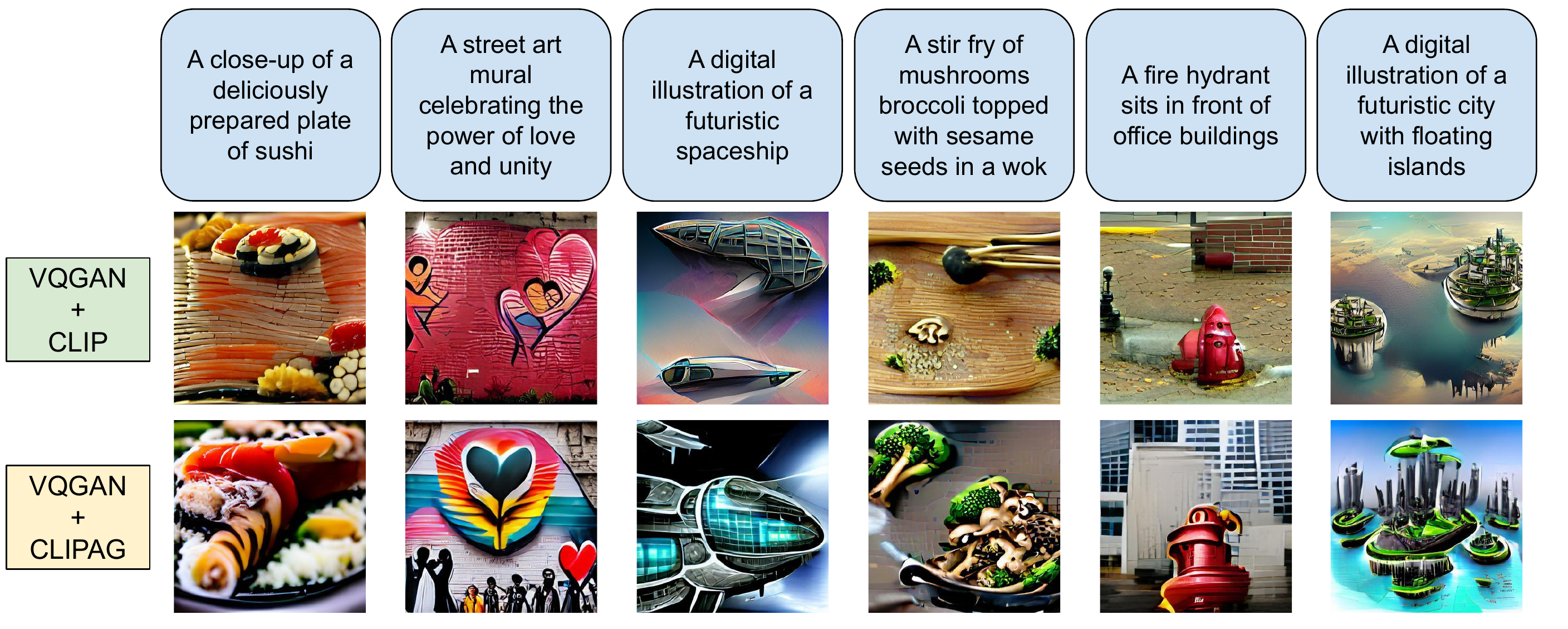}
    \caption{\textbf{VQGAN+CLIP results.} Visualization of VQAGAN + CLIP results using both CLIP and CLIPAG on various text prompts using the same hyperparameters and seed. As can be seen, seamlessly integrating CLIPAG to VQGAN+CLIP improves the generated images.}
    \label{fig:vqgan+clip}
    \vspace{-0.5cm}
\end{figure*}

Conducting such training brings several challenges and design choices.
First, CLIP was originally trained on massive $400$ million image-text pairs with a humongous batch size of $32,768$, using hundreds of GPUs. The combination of the training set size, which introduces a huge diversity, and the large mini-batches contributing to the effectiveness of contrastive training led to unprecedented capabilities. Finetuning such an architecture using academic resources might potentially lead to catastrophic forgetting~\cite{french1999catastrophic} and deteriorate performance. Moreover, when finetuning on adversarial examples, this could be further exacerbated. The vulnerability of CLIP to adversarial attacks might force a massive change of parameters during the finetuning, resulting in significant degradation of the generalization capabilities of CLIP, which are necessary for a wide range of downstream generative tasks.
However, recall that our objective is not to robustify CLIP but rather to align its gradients. Considering the challenges mentioned above, and the observation that in the unimodal case, adversarial training with even a low maximum perturbation bound can lead to perceptually aligned gradients~\cite{aggarwal2020benefits}, we focus on adversarial training using a small threat model. We hypothesize that this strategy will lead to PAG with less detrimental effects than applying adversarial training with the common robustification threat models.
In practice, we use a threat model of $\Delta = \{\delta \ : \ \lVert \delta \rVert_2 \leq 1.5\}$ for images of $224\times224\times3$, which is approximately equivalent to a mean pixel change of $\frac{1}{255}$, which is much smaller than the ones used in adversarial robustness research. 

In our approach, we focus on aligning the gradients of the image encoder, and thus, we choose to freeze the parameters of the pretrained text encoder when solving Equation \ref{eq:vl_adv_train}. By doing so, we not only reduce computational costs (freezing half of the model's parameters) but also introduce a valuable stabilizing mechanism during the adversarial finetuning of the image encoder. This is particularly effective because the pretrained text encoder generates meaningful representations, and aligning the image encoder with it can significantly enhance its performance. 
Furthermore, we conduct comprehensive experiments to thoroughly evaluate the impact of various design choices, which we discuss in the supplementary materials. Specifically, we investigate the effects of different architectures, including both convolutional networks and Vision Transformers (ViT)~\cite{dosovitskiy2021image}, as well as different threat models, such as $L_2$ and $L_\infty$ norms. In our practical implementation, we mainly focus on CLIP's {ViT-B/32} architecture, which is widely utilized in downstream tasks and thus enables a fair comparison.
We train it using a concatenated dataset that combines SBU~\cite{Ordonez:2011:im2text}, CC3M~\cite{sharma2018conceptual}, CC12M~\cite{changpinyo2021conceptual} and LAION-400M~\cite{schuhmann2021laion400m} and we subsample them to obtain a uniform dataset (\textit{i.e.}, the distribution to draw from each source is approximately the same). 
We perform a short finetuning with a low learning rate for the model using eight A40 GPUs while keeping the text encoder frozen (implementation details listed in \cref{sec:imp_det}).

After adversarially finetuning CLIP, we assess whether it achieves greater gradient alignment than the ``vanilla'' model and the randomized smoothed one. As illustrated in Figure~\ref{fig:VL_PAG}, the robustly trained variant exhibits the highest alignment with the provided texts. These findings indicate that similar to the unimodal case, robustification yields PAG in the context of VL models. Due to its superiority, we focus hereafter on the adversarially trained model.

\section{CLIPAG as a Generative Model}
\label{sec:experiments}
In this section, we explore the benefits of CLIPAG (CLIP with Perceptually Aligned Gradients) in various generative tasks in two main settings -- CLIP-based generative frameworks and generator-free text-to-image generation.

\subsection{Text-to-Image Generative Frameworks}
\label{sec:downstream}
Due to its strong vision-language alignment, CLIP is used as a fundamental block in various text-to-image generative tasks and applications, such as text-based editing~\cite{styleclip,stylegan-nada,text2live,clipasso,clipstyler} and generation~\cite{clipdraw,vqgan-clip,gigagan,glide,dalle2,latentdiffusion}.
In this section, we demonstrate that CLIPAG can be integrated into existing text-to-image generation applications in a ``plug-n-play'' manner by simply replacing the ``vanilla'' CLIP with its robust counterpart.
Specifically, to thoroughly explore the effects of leveraging CLIPAG, we experiment with it in text-based image editing and generation frameworks by integrating it into CLIPDraw~\cite{clipdraw}, VQGAN+CLIP~\cite{vqgan-clip} and CLIPStyler~\cite{clipstyler}. We focus mainly on CLIPDraw due to its simplicity and lack of a generative model, which enables us to explore the generative capabilities of CLIPAG compared to the standard CLIP.
In addition, in \cref{sec:explain}, we demonstrate that besides generative tasks, CLIPAG can lead to improved explainability.

\begin{figure}[t]
    \centering
    \includegraphics[width=\linewidth]{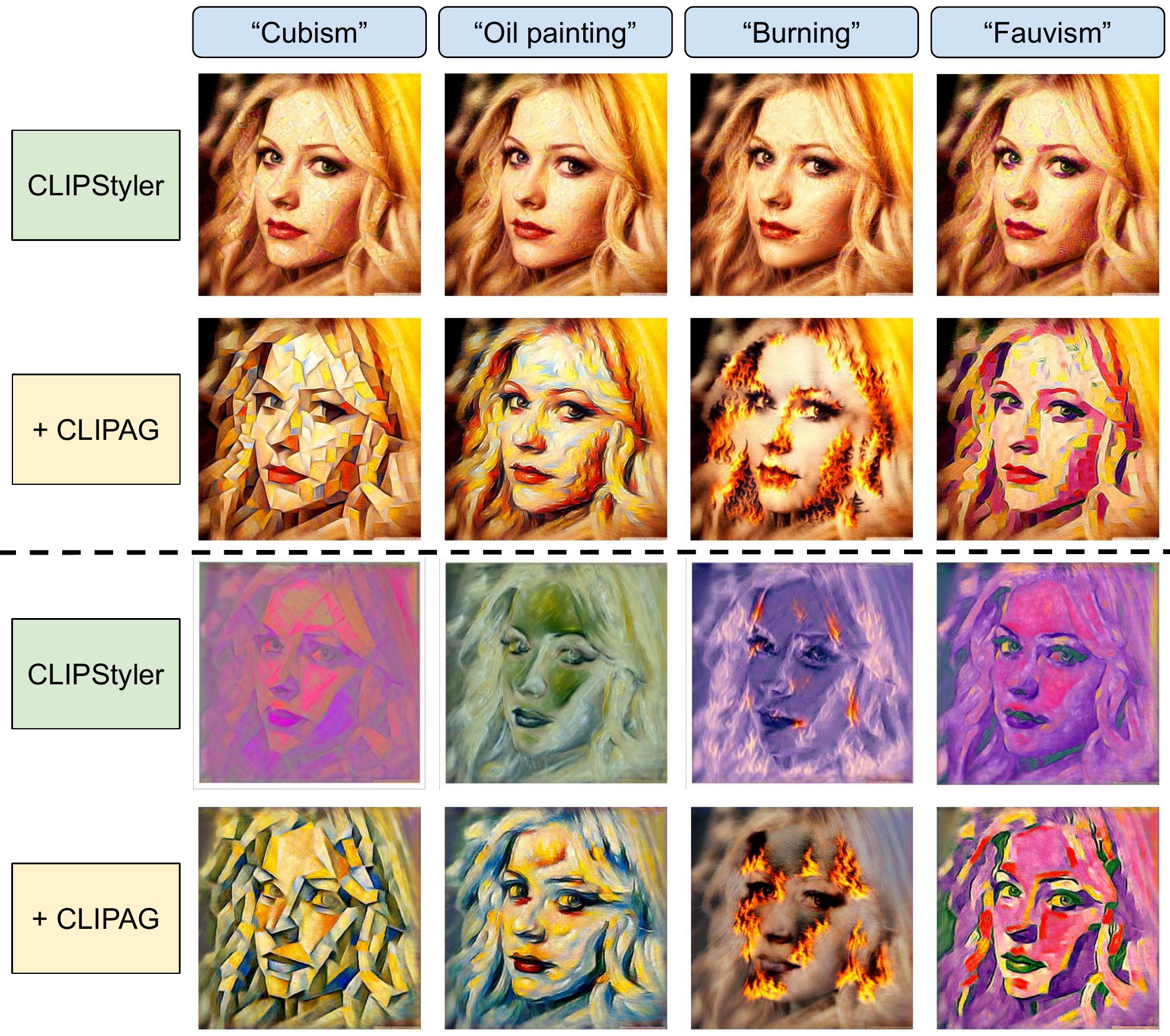}
    \caption{\textbf{CLIPStyler results.} Visualization of applying CLIPStyler~\cite{clipstyler} using both CLIP and CLIPAG, with and without the style network (bottom and top, respectively).
    As can be seen, while CLIP struggles to guide the style transfer process without a style network, CLIPAG leads to much better results, attesting to its improved gradients. Additionally, even with the introduction of the style network, the outputs of CLIPAG are more convincing.}
    \label{fig:clipstyler}
\end{figure}

\paragraph{Text-based Image Generation}
In this setting, we consider both CLIPDraw~\cite{clipdraw} and VQGAN+CLIP~\cite{vqgan-clip}.
CLIPDraw proposed a CLIP-based approach for text-to-drawing generation by optimizing a set of Bézier curves to minimize the cosine distance in the CLIP space between generated images and description prompts.
Additionally, to overcome the issue of the non-aligned gradients of CLIP, the authors utilized a multiview augmentation pipeline.
As stated in their paper, ``\emph{without image augmentation, synthesis through-optimization methods often result in adversarial images that fulfill the numerical objective but are unrecognizable to humans}.''
We study the effect of replacing CLIP in CLIPDraw with CLIPAG in two configurations -- with and without the augmentation pipe.
The rationale behind omitting the augmentation is to compare the generative capabilities of CLIP and CLIPAG.
First, we analyze the performance qualitatively and present the results in Figure~\ref{fig:clipdraw}. As can be seen, CLIPAG leads to improved performance compared to the baseline and is also capable of operating without augmentation, unlike the baseline, due to PAG.

 

\begin{table}[t]
 \setlength{\tabcolsep}{1.5pt}
 \centering
 \resizebox{\linewidth}{!} 
 {
 \begin{tabular}{|c c| c c |c c c c|}
 \hline
 \multirow{3}{*}{Method} & \multirow{3}{*}{Aug.} & \multicolumn{2}{c|}{\multirow{2}{*}{Aesthetics}} & \multicolumn{4}{c|}{Caption consistency} \\
 & & & & \multicolumn{2}{c}{ConvNext} & \multicolumn{2}{c|}{ViT-H/14} \\
 & & Score & Pref. & Sim. & R-Prec & Sim. & R-Prec \\
 \hline
 CLIPDraw~\cite{clipdraw} & \xmark & 3.70 & 22\% & 27.1 & 2\% & 18.1 & 1\%\\
 + CLIPAG & \xmark & \color{mygreen}{\textbf{3.98}} & \color{mygreen}{\textbf{78\%}} & \color{mygreen}{\textbf{32.1}} & \color{mygreen}{\textbf{19\%}} & \color{mygreen}{\textbf{26.3}} & \color{mygreen}{\textbf{32\%}} \\ \hline
 
 CLIPDraw~\cite{clipdraw} & \cmark & 4.14 & 34\% & 35.4 & 31\% & 31.6 & 70\%\\
 + CLIPAG & \cmark & \color{mygreen}{\textbf{4.31}} & \color{mygreen}{\textbf{66\%}} & \color{mygreen}{\textbf{36.0}} & \color{mygreen}{\textbf{52\%}} & \color{mygreen}{\textbf{32.0}} & \color{mygreen}{\textbf{72\%}} \\ \hline
 \end{tabular}
 }
 \caption{\textbf{CLIPDraw quantitative results.} CLIPDraw results using CLIP and CLIPAG using aesthetic and caption consistency metrics, with and without augmentation pipeline. As can be seen, simply replacing CLIP with CLIPAG leads to a substantial improvement in terms of aesthetics and caption similarity, attesting to the benefits of CLIP with Perceptually Aligned Gradients.}
 \label{tab:clipdraw}
 \vspace{-0.5cm}
\end{table}

In order to quantitatively analyze performance, we adopt an automatic procedure for aesthetics assessment.
To this end, we generate $100$ artistic prompts using ChatGPT~\cite{brown2020language} by conditioning on prompt examples from CLIPDraw's paper as a prefix. 
Next, we generate $100$ images using CLIPDraw with the baseline and CLIPAG.
Finally, we propose two main empirical metrics for assessing the performance -- (i) Utilize a linear-probed CLIP model on AVA dataset~\cite{murray2012ava}, a human-annotated dataset of aesthetics containing over $250,000$ images (the aesthetic scores are between 1 to 10). A similar technique was also adopted in ~\cite{dalle2}. Based on this model, we report the mean aesthetic score and the Preference rate between CLIPDraw using CLIP and CLIPAG. (ii) We utilize two publicly-available CLIP models~\cite{ilharco_gabriel_2021_5143773} (ConvNext\footnote{\href{https://huggingface.co/laion/CLIP-convnext_base_w_320-laion_aesthetic-s13B-b82K}{\text{CLIP-convnext-base-w-320-laion-aesthetic-s13B-b82K}}} and ViT-H/14\footnote{\href{https://huggingface.co/laion/CLIP-ViT-H-14-laion2B-s32B-b79K}{CLIP-ViT-H-14-laion2B-s32B-b79K}}) and report two metrics aimed at capturing the caption consistency -- the cosine similarity (Sim.) and the retrieval precision (R-Prec). R-Prec is averaged precision of a retrieval task where, for each generated image, the model predicts the most probable prompt among all the $100$ generated prompts. 
We report the above scores in Table~\ref{tab:clipdraw} both with and without augmentation.
As can be seen, CLIPAG significantly outperforms the baseline in all metrics when omitting the augmentations due to the PAG property.
Interestingly, even with the augmentations aimed at mitigating CLIP's susceptibility to adversarial attacks, CLIPAG leads to much-improved performance.

An additional evidence of the benefits of CLIPAG in text-based generative frameworks is provided by using VQGAN+CLIP~\cite{vqgan-clip}.
In Figure~\ref{fig:vqgan+clip}, we qualitatively demonstrate the effectiveness of replacing CLIP with CLIPAG in the VQGAN+CLIP framework on different prompts.
To quantitatively evaluate the effect of CLIPAG in the VQGAN+CLIP framework, we randomly sample $100$ captions from the validation set of MS-COCO captions~\cite{chen2015microsoft}, generate images accordingly with both CLIP and CLIPAG and calculate their cosine similarity using OpenCLIP {ViT-H/14}.
Despite the fact that OpenCLIP is not robust and probably more aligned with the ``vanilla'' one, CLIPAG obtains {\color{mygreen}{$\mathbf{34.3}$}} cosine similarity, surpassing CLIP's {\color{red}{$\mathbf{33.4}$}}.
We provide additional details regarding these experiments in \cref{sec:imp_det}, along with additional qualitative results.

\paragraph{Text-based Image Editing}
For image editing tasks, we adopt CLIPStyler~\cite{clipstyler} as our framework, which leverages a pretrained CLIP model for text-based image style transfer. CLIPStyler addresses the issue of meaningless CLIP gradients by incorporating a style network and employing a multiview augmentation pipeline.
In this study, we compare the performance of CLIPStyler using the standard CLIP and CLIPAG model in two settings: (i) the ``plug-n-play'' approach and (ii) without the style network (to isolate the guidance capability of CLIPAG and CLIP in style-transfer). As can be seen in Figure~\ref{fig:clipstyler}, seamlessly replacing CLIP with CLIPAG leads to marked improvement. In addition, while utilizing CLIP without the style network completely fails, CLIPAG's results are substantially better, showcasing its capabilities in guiding a text-based style transfer process.

 

\subsection{Generator-Free Text-to-Image Generation}
\label{sec:gen}
\begin{figure}[t]
    \centering
    \includegraphics[width=\linewidth]{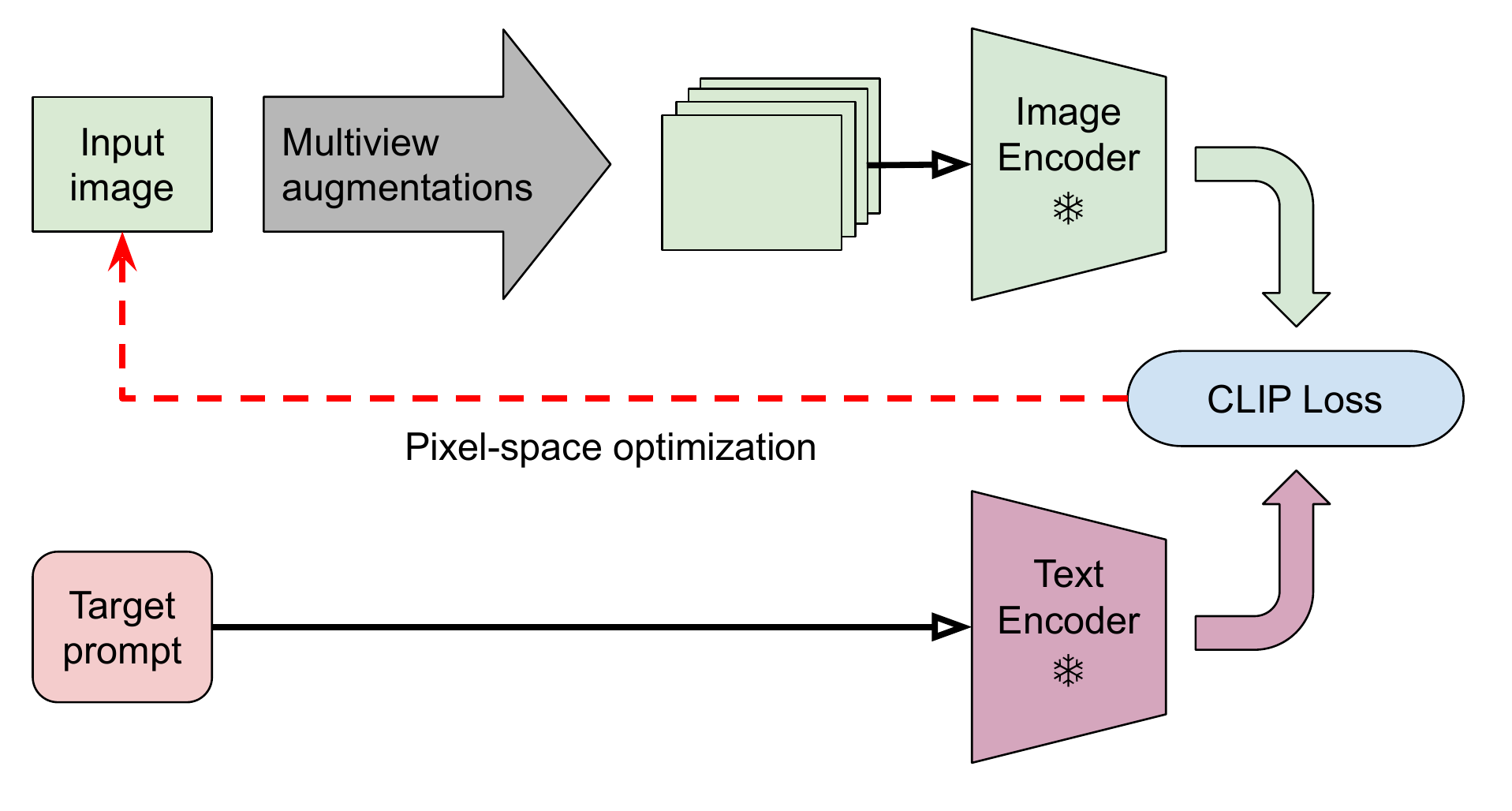}
    \caption{\textbf{Generator-free text-to-image framework.} Visualization of the proposed method to harness CLIPAG for text-to-image generation without any generative model. Specifically, we iteratively update the pixels of an image to better align to a textual description. Due to PAG, such a process leads to meaningful images that align with the text.}
    \label{fig:generation}
    \vspace{-0.5cm}
\end{figure}

In recent years, text-to-image generation has gained substantial attention, leading to the development of various methods aimed at tackling this challenging task~\cite{vqgan-clip, ramesh2021zeroshot, dalle2, saharia2022photorealistic, li2023blipdiffusion, rombach2022highresolution}. While these methods employ diverse techniques, they all rely on powerful generative models. In this section, we present a novel approach that breaks away from traditional text-to-image pipelines by leveraging CLIPAG, a compact, non-generative model with approximately $150$ million parameters.
To this end, we introduce a simple yet effective \emph{generator-free text-to-image synthesis} framework, as depicted in Figure~\ref{fig:generation}, which employs an iterative pixel-space optimization technique to align an input image with a target text, leveraging the Perceptually Aligned Gradients of CLIPAG. Notably, our approach does not involve training any of CLIPAG's components.
The proposed framework consists of three key blocks: 
(i) \textbf{Initialization} -- Sampling the input for the generation process. To achieve this, we model a low-resolution dataset as a Gaussian Mixture Model and select the candidate image with the best alignment to the given text based on CLIP; (ii) \textbf{Multiview augmentations} -- in each iteration, we duplicate the image and perform differential random augmentations. In practice, we use DiffAugment~\cite{zhao2020differentiable} and random cropping; (iii) \textbf{CLIPAG-based Loss} -- in every step, we update the input image using the input gradients of CLIPAG to better align with the text.
We provide the implementation details in \cref{sec:imp_det}.

To qualitatively demonstrate the capabilities of our generator-free framework, we visualize generated images corresponding to various textual descriptions in Figure~\ref{fig:generationExamples} and Appendix~\ref{sec:app_gen_free}. The images produced by CLIPAG exhibit a high level of visual fidelity and consistency with the given text, despite the absence of a traditional generative model.
For quantitative assessment, we conduct experiments on the MS-COCO dataset. First, we synthesize the same $100$ prompts used in the VQGAN+CLIP experiments and calculate the CLIPScore using OpenCLIP ViT-14/H. Our framework achieves an impressive score of {\textcolor{mygreen}{$\mathbf{36.8}$}}, surpassing the performance of the generator-based framework. Specifically, we outperform VQGAN with CLIP by $\textcolor{mygreen}{\uparrow \mathbf{3.4}}$ and VQGAN with CLIPAG by $\textcolor{mygreen}{\uparrow \mathbf{2.5}}$.
Next, to further evaluate the performance, we generate 30,000 images from MS-COCO validation captions in a zero-shot (ZS) setting. We calculate the CLIPScore~\cite{hessel2022clipscore}, Inception Score (IS)~\cite{salimans2016improved}, and Frechet Inception Distance (FID)~\cite{heusel2018gans}, and compare our results  against strong baselines in Table~\ref{tab:gen}.
Notably, while using a significantly smaller non-generative model, our approach outperforms DALL-E~\cite{ramesh2021zeroshot} and CogView~\cite{ding2021cogview} in Inception Score. However, our FID metric is relatively weaker, potentially attributed to CLIPAG's tendency to generate colorful and artistic images, which deviate from the characteristics of the MS-COCO. 
We hypothesize that this can be mitigated via prompt tuning but leave this for future work.
In addition, we explore different aspects of our proposed scheme, in \Cref{sec:app_gen_free}.
While our results do not yet rival state-of-the-art methods~\cite{ramesh2022hierarchical, saharia2022photorealistic, glide}, they highlight the remarkable generative capabilities of CLIPAG and potentially pave the way for a new family of generator-free text-to-image generation techniques.

\begin{table}[t]
    \setlength{\tabcolsep}{5pt}
    \resizebox{\linewidth}{!} 
    {
    \centering
    \begin{tabular}{|l|c|c|c|c|c|}
        \hline
        Method & \#Params. & ZS & IS$\uparrow$ & FID$\downarrow$ & CLIPScore$\uparrow$ \\
        \hline
        Stack-GAN~\cite{zhang2018stackgan} & - & \xmark & 8.5 & 74.1 & - \\
        AttnGAN~\cite{xu2018attngan} & 230M & \xmark & \textbf{23.3} & 35.5 & 27.7 \\
        CogView~\cite{ding2021cogview} & 4,000M & \xmark & 18.2 & \underline{27.1} & \underline{33.2} \\
        DALL-E~\cite{ramesh2021zeroshot} & 12,000M & \cmark & 17.9 & 27.5 & - \\
        GLIDE~\cite{glide} & 6,000M & \cmark & - & \textbf{12.2} & -\\
        \hline
        Ours & 150M & \cmark & \underline{18.7} & 42.3 & \textbf{34.7} \\
        \hline
    \end{tabular}
    }
    \caption{\textbf{MS-COCO text-to-image generation results}. Inception score, CLIPScore (higher is better), and Frechet Inception Distance (lower is better) results, along with model sizes.}
    \label{tab:gen}
    \vspace{-0.5cm}
\end{table}

\vspace{-0.2cm}
\section{Discussion and Conclusions}
\label{sec:disc}
In this paper, we explore the concept of PAG in the context of Vision-Language architectures using CLIP. Our findings highlight several significant contributions. First, we establish the presence of PAG in CLIP by adversarially finetuning it. This demonstrates that the phenomenon of PAG is not limited to unimodal vision-only architectures but extends to multimodal models. 
Second, we demonstrate that CLIPAG can be seamlessly integrated into existing text-to-image existing frameworks, leading to substantial improvements. 
Lastly, we showcase that CLIPAG can be used for generator-free text-to-image synthesis, which typically relies heavily on generative models. Our results demonstrate the practical implications and potential of harnessing PAG in real-world Vision-Language applications. We believe the insights and findings presented in this paper will inspire further exploration and advancements in harnessing PAG in multimodal research.


{\small
\bibliographystyle{ieee_fullname}
\bibliography{egbib}
}

\clearpage

\appendix
\section{Implementation Details}
\label{sec:imp_det}
\subsection{Training}
We implement our code based on the publicly-available CLIP repository of OpenCLIP~\cite{ilharco_gabriel_2021_5143773}\footnote{\url{https://github.com/mlfoundations/open_clip}}.
In all the experiments considered in the paper, we use a pretrained {ViT-B/32} architecture, initialized using OpenCLIP's weights\footnote{\url{https://huggingface.co/laion/CLIP-ViT-B-32-laion2B-s34B-b79K}}.
In addition, we consider a threat model of $L_2$ with a maximum perturbation of $1.5$ and $5$ PGD steps by extending the standard implementation to the multimodal case.
We set the learning rate and weight decay to $2e-5$ and $1e-4$, respectively, and perform $10$ gradient accumulation steps to enable larger batch sizes, resulting in an effective batch size of $40,960$.
As for the training data, we concatenate SBU, CC-3M, and CC-12M without resampling and a downsampled version of LAION 400M (by $\times0.04$).
We freeze the textual encoder and finetune the vision encoder ($88$M parameters) on eight A40 GPUs.
We study the effects of different design choices in \cref{sec:ablation}.
We will make our code and pretrained model publicly available.

\subsection{Text-to-Image Generative Frameworks}
In the experiments presented in the main paper, we seamlessly replace the existing CLIP {ViT-B/32} in such frameworks with CLIPAG, using the same architecture and hyperparameters as in the baseline.
In this way, we ensure a fair comparison that enables us to study the benefits of CLIPAG.
We explore our approach in three main frameworks and describe implementation details and relevant information below.

\paragraph{CLIPDraw}
We utilize the official implementation of CLIPDraw\footnote{\url{https://colab.research.google.com/github/kvfrans/clipdraw/blob/main/clipdraw.ipynb}} and replace the used ``vanilla'' CLIP {ViT-B/32} by CLIPAG with the same architecture.
As stated in the main paper, we experiment with two settings -- with and without augmentations.
Besides the qualitative demonstrations, we propose a quantitative evaluation procedure to evaluate the performance.
To this end, we utilize some prompts suggested in CLIPDraw's paper and request from ChatGPT~\cite{brown2020language} to provide us additional 100 similar prompts.
We synthesize the generated prompts using CLIPDraw with CLIP and CLIPAG, with and without augmentations. Next, we calculate the aesthetic score using a CLIP trained on human aesthetic predictions using a publicly available code\footnote{\url{https://github.com/christophschuhmann/improved-aesthetic-predictor}}. 
Given an image, such a model outputs a continuous value describing the aesthetic score (higher is better).
Using this model, we calculate an aesthetic score for every generated drawing and report two metrics -- Average aesthetics score and a pairwise aesthetic preference.
Since the proposed aesthetic metrics do not depend on the caption, we also measure caption similarity using CLIP similarity with two CLIP models to validate the results better.
Besides CLIP similarity, we utilize the R-Prec metric~\cite{jain2022vectorfusion}, focusing on image-based text retrieval. Specifically, given a generated image, we use CLIP to pick the most probable prompt across all the 100 textual descriptions. We average the accuracy of such a task for the generated drawings, resulting in the R-Prec metric, similar to~\cite{jain2022vectorfusion}.
The combination of these metrics captures both the generated drawings' quality and consistency with the caption, enabling a proper evaluation of the generated drawings.

\begin{figure*}[t!]
    \includegraphics[width=\textwidth]{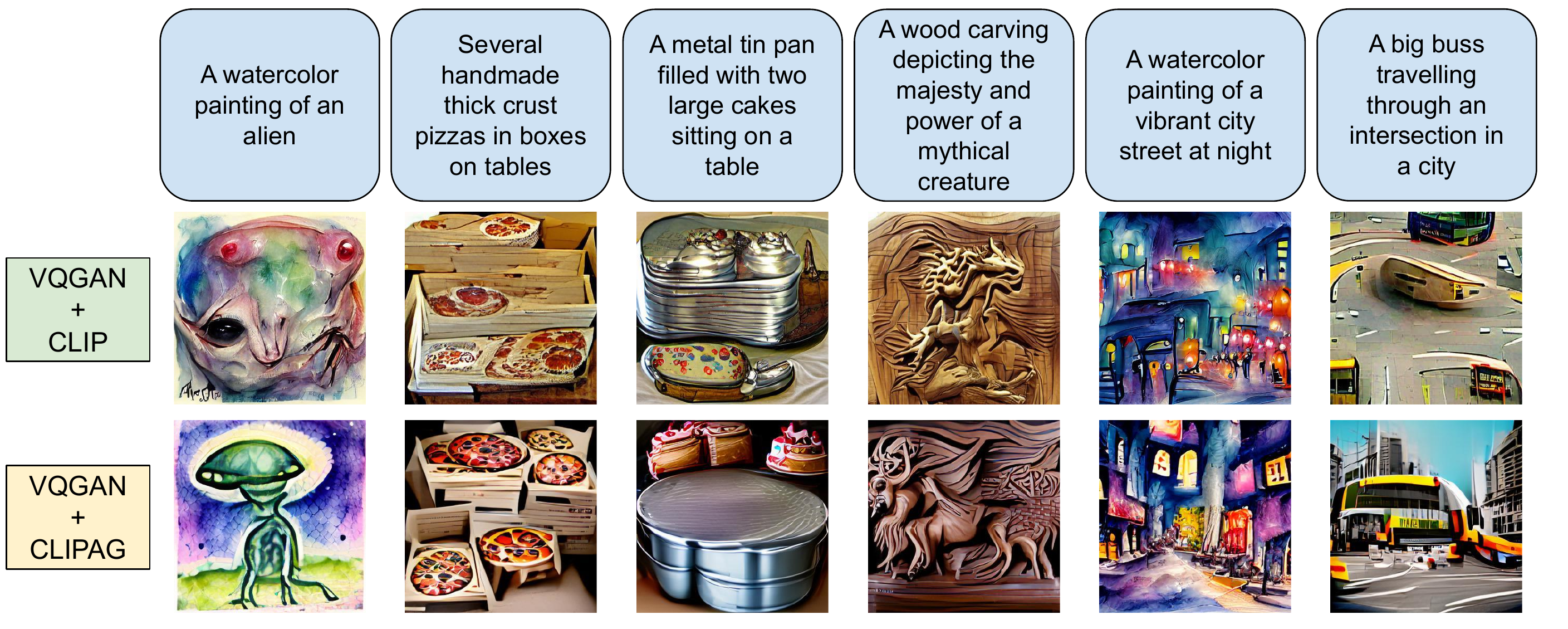}
    \caption{\textbf{VQGAN+CLIP additional results.}}
    \label{fig:vqgan+clip_app}
\end{figure*}

\paragraph{VQGAN+CLIP}
Similar to CLIPDraw, we use the official code repository~\footnote{\url{https://github.com/nerdyrodent/VQGAN-CLIP}} and replace CLIP with CLIPAG.
We randomly sample 100 captions from the validation set of the MS-COCO dataset and generate two sets of 100 images. Next, we calculate the CLIP similarity to measure the alignment of the generated images with the desired prompts.
To better demonstrate the effects of replacing CLIP with CLIPAG, we provide additional results in~\cref{fig:vqgan+clip_app}.

\paragraph{CLIPStyler}

\begin{figure}[t!]
    \includegraphics[width=\linewidth]{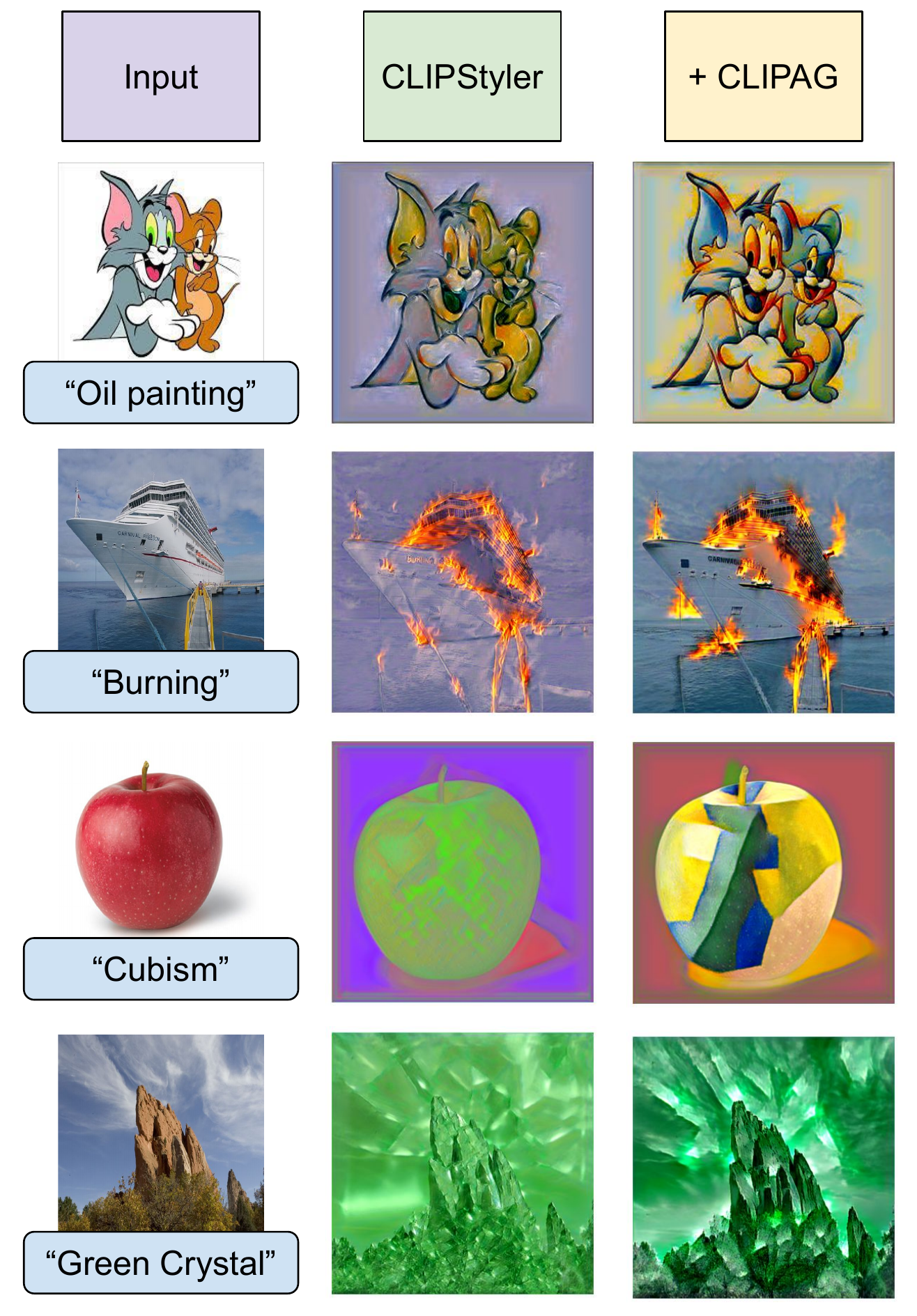}
    \caption{\textbf{CLIPStyler additional results.} Style transfer results of multiple images using different textual prompts with CLIP and CLIPAG while using the style network.}
    \label{fig:clipstyler_app}
\end{figure}

We experiment with CLIPStyler official code\footnote{\url{https://github.com/cyclomon/CLIPstyler}} and replace CLIP with CLIPAG, with and without the style network. 
To better demonstrate the effectiveness of CLIPAG in the text-guided style transfer context, we provide additional results in ~\cref{fig:clipstyler_app}.

\begin{figure}[t!]
    \includegraphics[width=\linewidth]{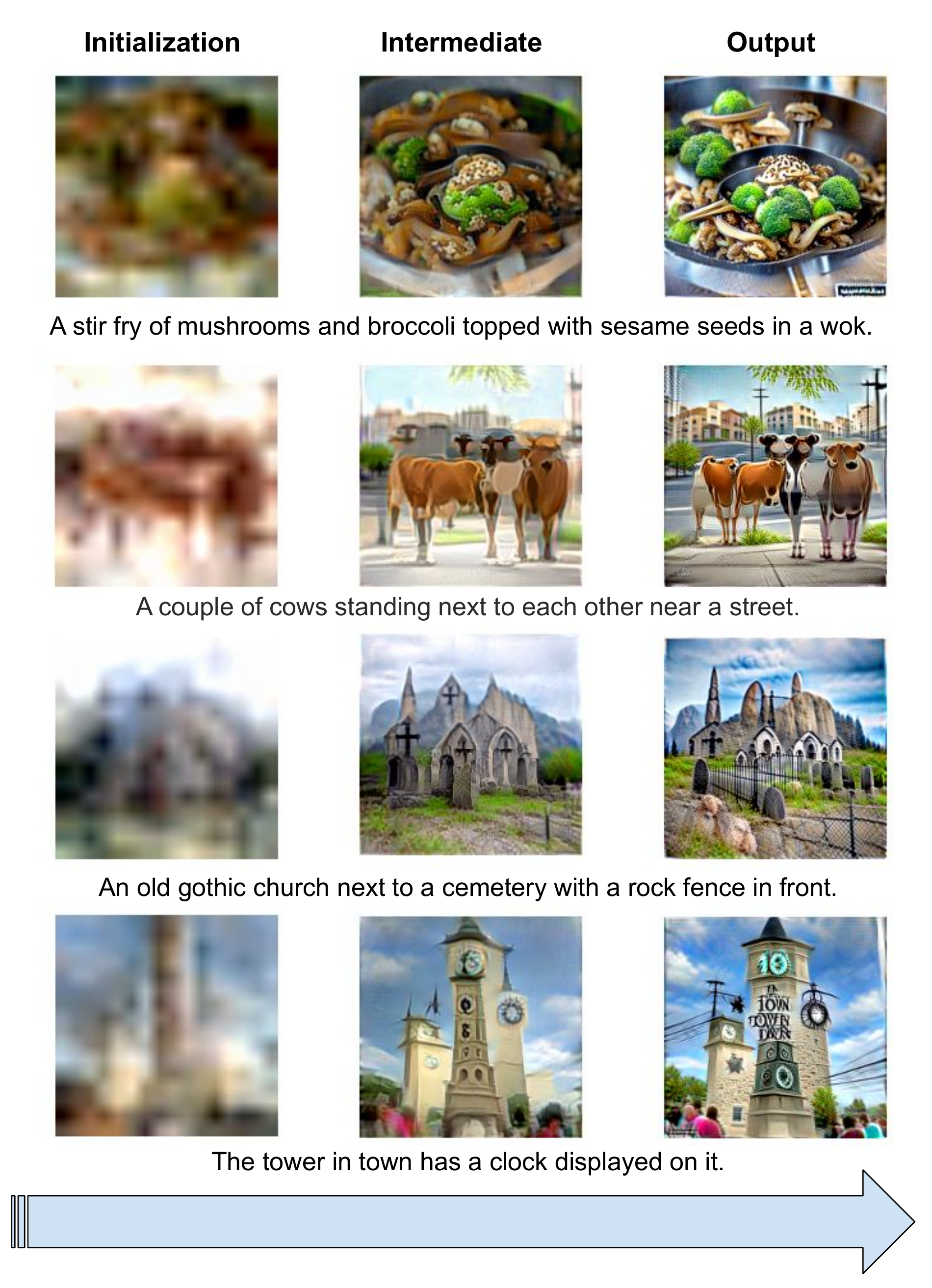}
    \caption{\textbf{Generation Trajectory}. Visualization of three different time steps in the generation process.}
    \label{fig:gen_trajectory}
\end{figure}

\subsection{Generator-Free Text-to-Image Generation}
\begin{figure*}
    \centering
    \includegraphics[width=\textwidth]{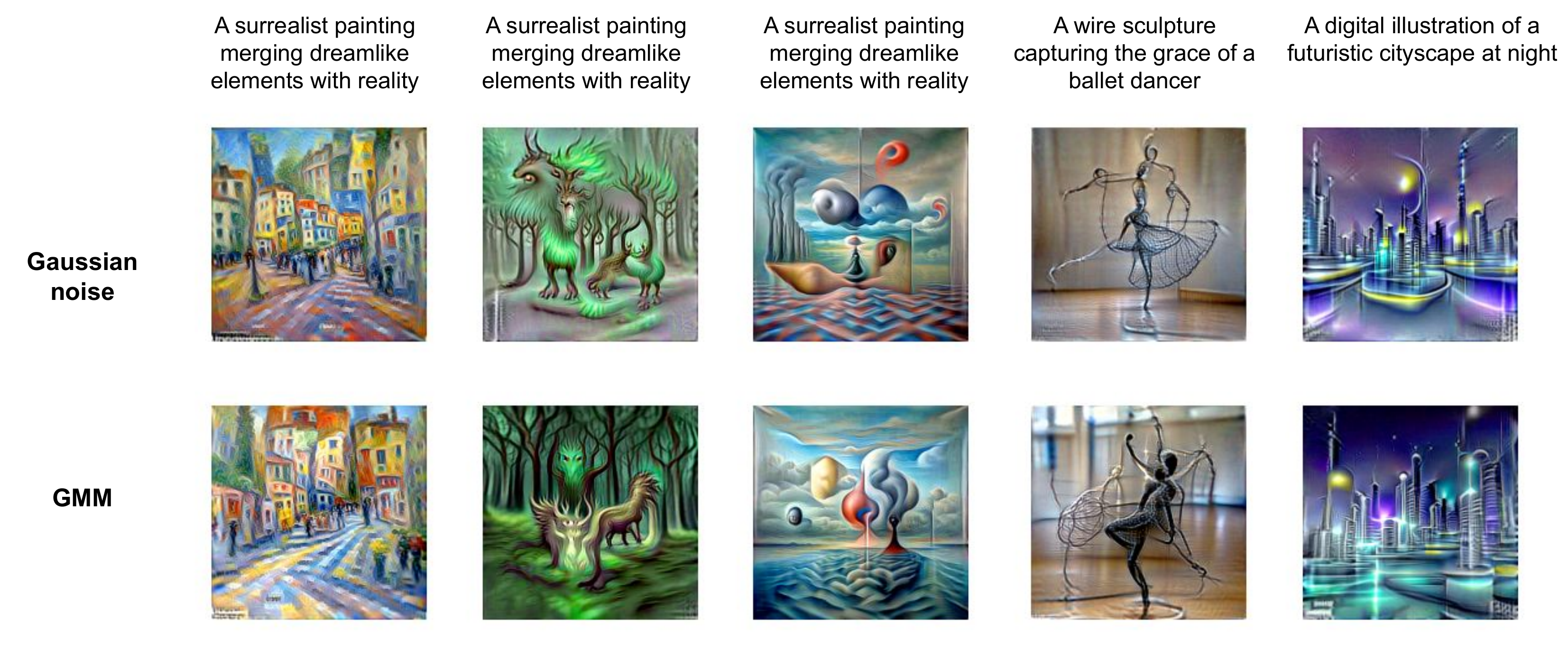}
    \caption{\textbf{Initialization ablation.} Comparison of our GMM initialization with random Gaussian noise in text-to-image generation.}
    \label{fig:init}
\end{figure*}

\paragraph{Initialization mechanism}
In practice, we propose the following simple-yet-effective initialization process -- we randomly sample image candidates from a simple distribution and pick the one that best matches the target text using CLIP cosine similarity.
Specifically, we leverage a downsampled version ($16\times16$) of the Tiny-ImageNet dataset~\cite{wu2017tiny} and train a Gaussian Mixture Model where each Gaussian represents a class. Next, we sample $M$ candidates from each of the $200$ classes, resulting in $M\times200$ images. 
Such images are unrealistic and mainly contain colorful blobs (a visualization of the chosen initial images is shown in Figure~\ref{fig:gen_trajectory}).
Next, we upsample these images to $224\times224$ and pick the one with the highest alignment with the target text as the input to our process, resulting in a generated image.
To study the effect of the initialization mechanism, we generate several prompts using the above-described initialization, compared to random Gaussian noise one in ~\Cref{fig:init}.
As can be seen, CLIPAG is capable of producing meaningful results with both initializations, attesting to its guiding capabilities. 

\paragraph{The generation process}
After the initialization step, we perform an iterative process of $K$ steps (set empirically to $1000$) in which we modify the input image to better align with the given textual description.
Specifically, we duplicate the image and augment every instance using random augmentations, leading to a batch of different image views.
Next, we input the batch to the image encoder to obtain feature representations.
Finally, we calculate the cosine similarity loss, calculate the input gradients and use them to update the image.
Unlike other works that harness CLIP~\cite{clipstyler,styleclip,stylegan-nada,text2live}, we do not use additional losses such as direction-loss and total variation regularization to guide the process but rather focus solely on the basic CLIP-loss.
Repeating these steps $K$ times results in pleasing generated images corresponding to the target captions.

\section{Explainability}
\label{sec:explain}
\begin{figure}[t]
    \centering
    \includegraphics[width=\linewidth]{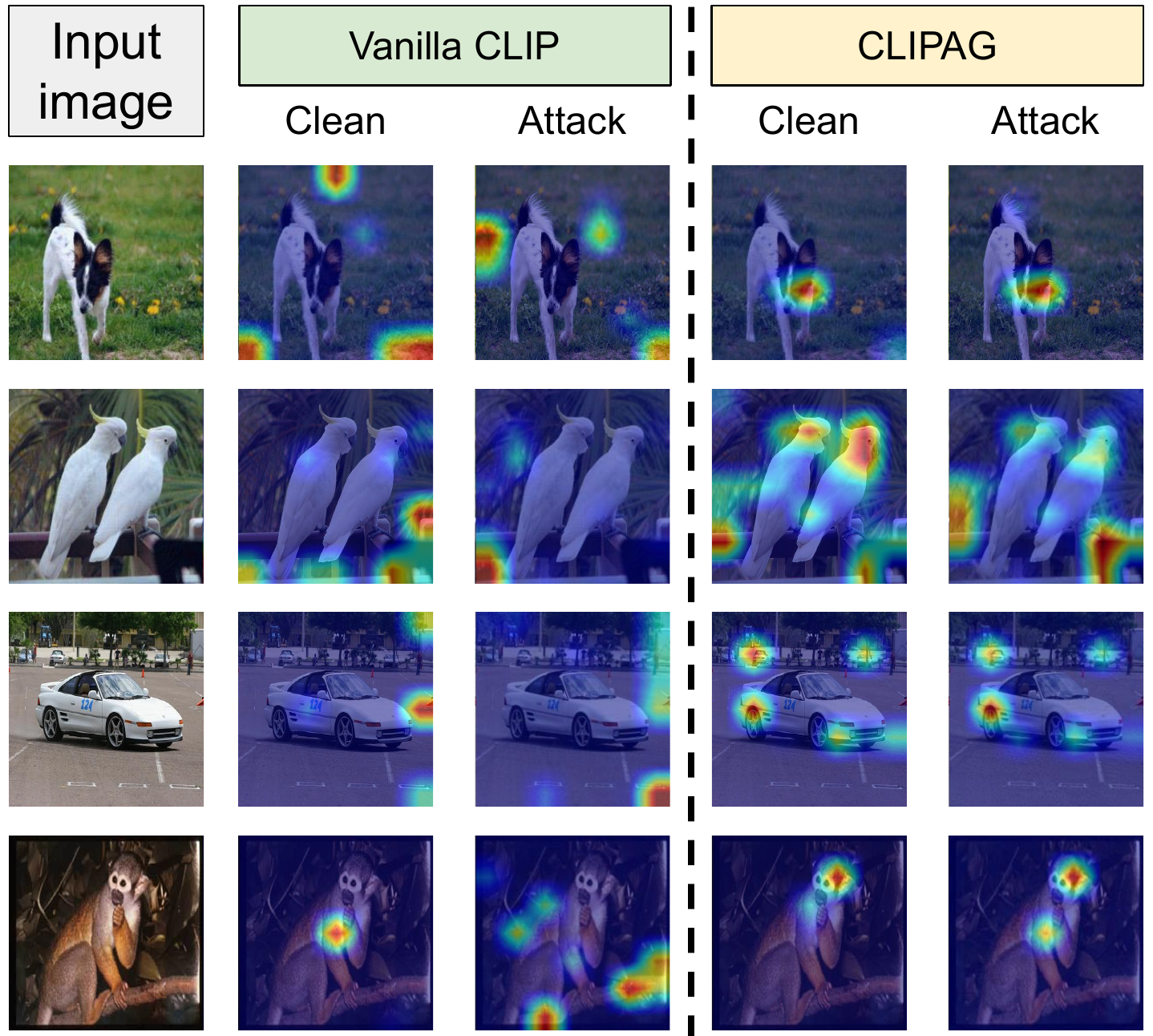}
    \caption{\textbf{Explainability visualizations.} GradCAM~\cite{gradcam} heatmaps for both the baseline CLIP ViT-B-32 and CLIPAG on ImageNet images. The targets for the GradCAM are \texttt{dog, parrot, cars} and \texttt{monkey}, respectively. As can be seen, both in the clean and adversarial cases, CLIPAG heatmaps are more aligned with the objects, providing better interpretability.}
    \label{fig:explain}
\end{figure}

With the introduction of learning-based machines into ``real-world'' applications, the interest in interpreting the decisions of such models has become a central concern.
Thus, the explainability of deep learning-based models is a crucial objective for improving the trust and transparency of such models. 
Moreover, it enables users to understand model predictions better and detect shortcuts and biases.
We hypothesize that due to its more aligned gradients, CLIPAG possesses improved explainability capabilities than the regular CLIP model.
To verify if this is indeed the case, we utilize The GradCAM~\cite{gradcam} (Gradient-weighted Class Activation Mapping) algorithm, which utilizes the model's gradients to generate visual heatmaps, highlighting the important regions in an input image for a given target.
We follow the implementation of ~\cite{chen2022gScoreCAM}\footnote{\url{https://github.com/anguyen8/gScoreCAM}}.
Specifically, GradCAM combines the features and the gradients
of a network’s layer by multiplying it. As this method relies on the gradients of the deepest layers, upsampling its results to the input resolution often leads to coarse results.
In addition, GradCAM is designed for convolutional neural networks and is significantly less effective in vision transformers.
In Figure~\ref{fig:explain}, we present the results of applying GradCAM on CLIP (using {ViT-B/32}) with both the original and CLIPAG, using ImageNet images in a zero-shot setting.
As can be seen, while GradCAM performs unsatisfactorily on the regular CLIP, applying it on CLIPAG leads to more aligned heatmaps with the target objects. 
We hypothesize that this improvement stems from the Perceptually Aligned Gradients property of CLIPAG, leading to an improved explainability with GradCAM.

Furthermore, we study the interpretability of CLIPAG under adversarial attacks. To this end, given an input image $\mathbf{x}$ and a textual description of an object $t$ (\textit{e.g.}, ``\texttt{a cat}''), we perform adversarial attacks to minimize the cosine similarity between $\textbf{x}$ and $y$ in the feature space, and maximize the one between the image and the negation of the textual description $t$, denoted as $\widetilde{t}$ (\textit{e.g.}, ``\texttt{not a cat}'').
Formally, we solve the following optimization problem:
\begin{equation}
    \max_{\delta\in\Delta}\mathcal{L}_{SIM} (f_{\theta_I}^I(\mathbf{x} + \delta), f_{\theta_T}^T(t)) - \mathcal{L}_{SIM} (f_{\theta_I}^I(\mathbf{x} + \delta), f_{\theta_T}^T(\widetilde{t}))
\end{equation}
where $\mathcal{L}_{SIM}$ is the cosine similarity loss, \textit{i.e.}, maximizing it minimizes the cosine similarity.
In particular, we perform a Projected Gradient Descent (PGD) attack where $\Delta=\{\ \delta : \lVert \delta \rVert_{\infty} \leq \frac{8}{255} \}$ using 20 steps and a step size of $\frac{1}{255}$. We provide visualizations of the outputs of GradCAM using both CLIPAG and the ``vanilla'' CLIP on adversarial attacks in Figure~\ref{fig:explain}. 
As can be seen, the adversarial attacks change the outputs of the baseline significantly; however, CLIPG's outputs are much more robust.


\section{Ablation Study}
\label{sec:ablation}
\begin{figure*}[t!]
    \centering
    \includegraphics[width=\textwidth]{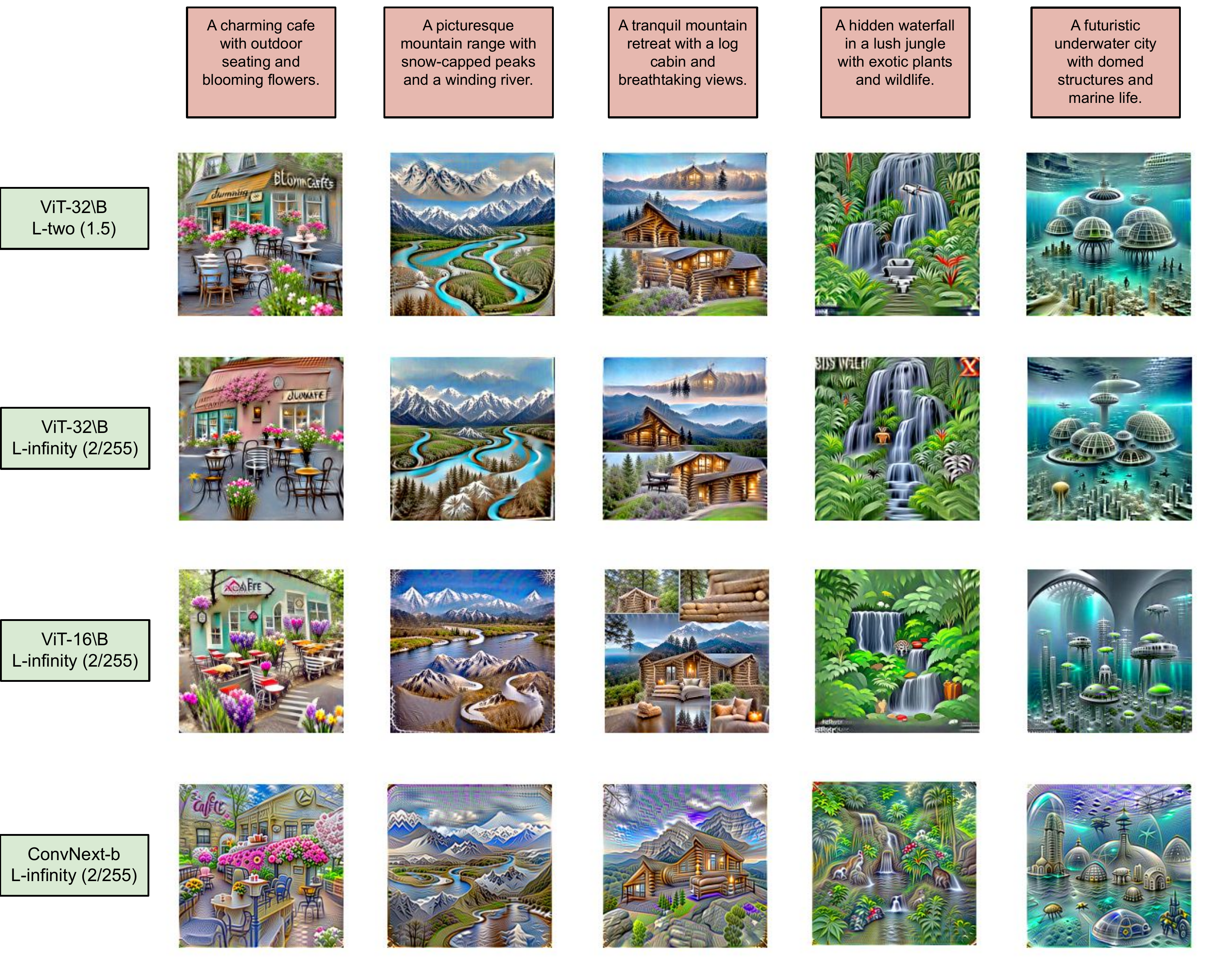}
    \caption{\textbf{CLIPAG generator-free text-to-image ablation study}.
    The effect of different design choices in the context of generator-free image synthesis, considering different architectures and threat models.}
    \label{fig:design_choices}
\end{figure*}

In this section, we explore the effects of different design choices of Vision-Language adversarial finetuning CLIP on its Perceptually Aligned Gradients.
To this end, we conduct a relatively short training and study the effect of different architectures ({ViT-B/32}, {ViT-B/16}, and {ConvNext}).
Moreover, we also compare an $L_\infty$ ($\epsilon=\frac{2}{255}$) to an $L_2$-based one ($\epsilon=1.5$).
To assess the impact of such design choices, we use them for generator-free text-to-image generation using our proposed framework, described in \Cref{sec:gen}, and visualize the results in \cref{fig:design_choices}.
As can be seen, all the different design choices lead to satisfactory outputs, attesting to their PAG.
However, there are some differences:
\begin{itemize}
    \item \textbf{Different ViTs} -- We consider both {ViT-B/16} and {ViT-B/16}, which differ in the patch size, using $L_\infty$-based threat model. The {ViT-B/16}, which utilizes a smaller path size, leads to some visual artifacts. We hypothesize that maximizing the consistency with the caption with a small patch architecture leads to fine-grained modifications that result in undesired artifacts.
    \item \textbf{CNN vs. ViT} -- While both are trained on the same threat model, the {ConvNext} guides the generation process towards images with significantly more saturated colors than the ViTs.
    \item \textbf{Threat model} -- We compare the $L_2, \ \epsilon=1.5$ to $L_\infty, \ \epsilon=\frac{2}{255}$ using {ViT-B/32}. Interestingly, despite these threat models being substantially different, they both lead to generated images with similar characteristics. Nevertheless, we find the results of the $L_2$ case more pleasing.
\end{itemize}
\noindent Thus, we mainly focus on the {ViT-B/32} architecture and the $L_2$ threat model.

\section{Generator-Free Text-To-Image Analysis}
\label{sec:app_gen_free}
In this section, we provide additional information and study different aspects of our text-to-image generation results in the generator-free setting, presented in Figure~\ref{fig:generation}.
According to our procedure, we perform an iterative process of updating $K$ steps (empirically set to $1000$).
During such updates, the generated image is modified to be more aligned with the textual description, according to CLIPAG.
To better study the effect of the iterative process, we depict samples in three timesteps in Figure~\ref{fig:gen_trajectory}-- (i) the initialization image, (ii) an intermediate image, and (iii) the final result.
In particular, \texttt{Initialization} depicts the starting point of our iterative process, \textit{i.e.}, a sample from our GMM that best aligns with the given text.
As can be seen in the Figure, the starting point is not a real image, as modeling images via GMMs is limited due to their high dimensionality.
Interestingly, CLIPAG is capable of transforming such inputs into perceptually meaningful content that corresponds with the text.
One can see that in the \texttt{intermediate} point, the resulting images contain most of the high-level features.
The process from the intermediate point towards the \texttt{output} adds mainly low-level details and refines the visual content. 

\begin{figure*}[t!]
    \centering
    \includegraphics[width=0.95\textwidth, height=0.95\textheight,keepaspectratio]{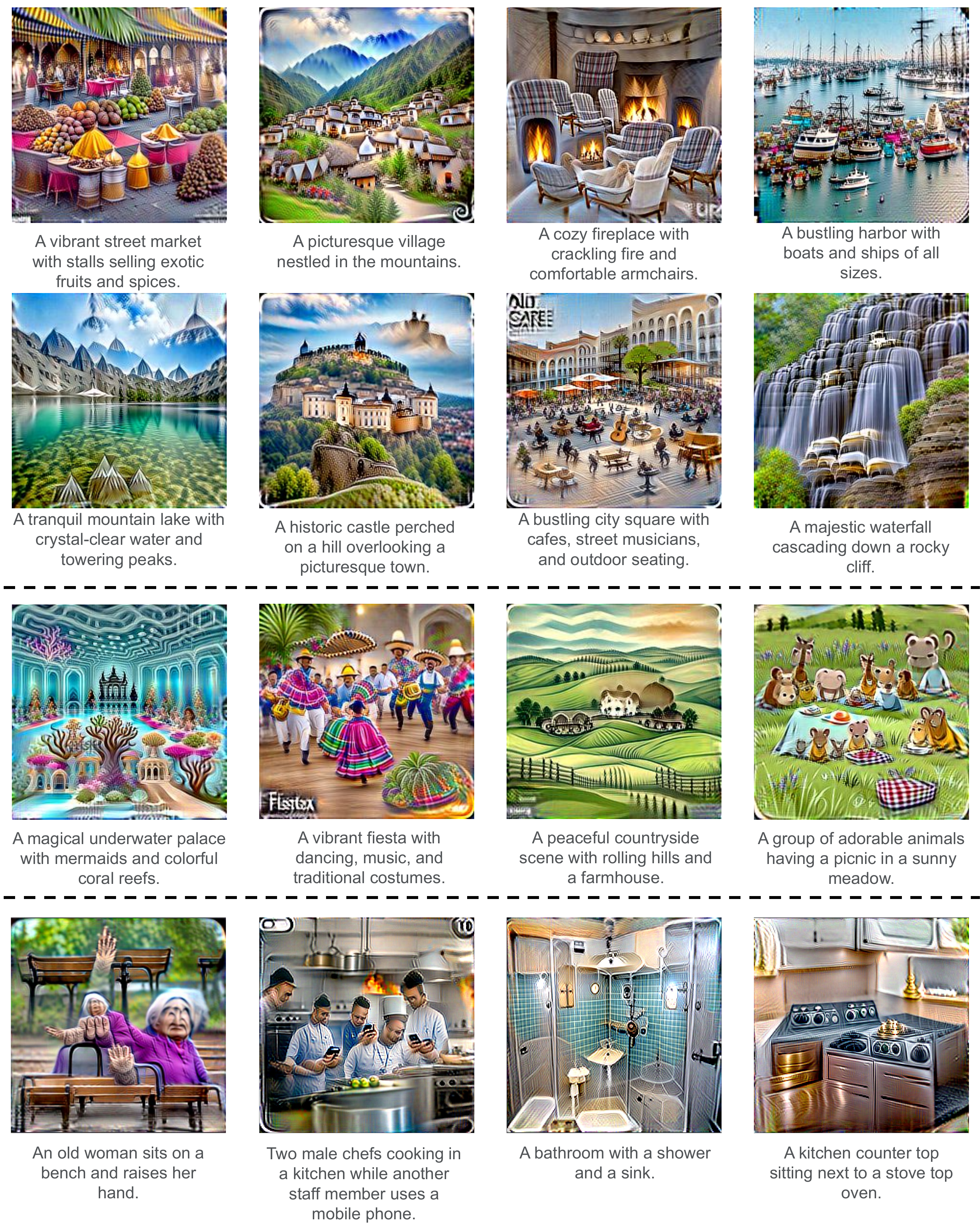}
    \caption{\textbf{CLIPAG generator-free text-to-image additional results}.
    The top two rows present additional generator-free synthesis results using CLIPAG. The third row demonstrates a phenomenon in which CLIPAG often opts for cartoonish and artistic content rather than a realistic one. In the last row, we depict some fail cases in which the resulting images are inconsistent.}
    \label{fig:gen_app}
\end{figure*}

To better understand the generative capabilities and explore different trends in the synthesis process, we provide additional qualitative results in Figure~\ref{fig:gen_app}.
As can be seen, in the top two rows, the outputs of our proposed algorithm are relatively natural and realistic.
However, CLIPAG often prefers ``cartoonish'' outputs over realistic ones, as can be seen in the third row.
We hypothesize that this might be affected by certain words in the target text prompt that guides the model towards such outputs (\textit{e.g.,} ``\texttt{magical}'').
We suspect such a tendency leads to lower FID scores when measured w.r.t. natural images dataset, such as MS-COCO.
In addition, our model sometimes maximizes the alignment with the given text by producing OCR~\cite{DBLP:journals/corr/abs-2205-03873,Litman_2020_CVPR} information that corresponds with the caption.
For example, in the third row in Figure~\ref{fig:gen_app}, the model attempts to spell the word ``\texttt{fiesta}'', which appears in the caption.
Similarly, in the bottom row of Figure~\ref{fig:gen_trajectory}, the model spells ``\texttt{town}'' and ``\texttt{tower}'' that are included in the caption.

\begin{figure*}[t]
    \centering
    \includegraphics[width=0.8\textwidth]{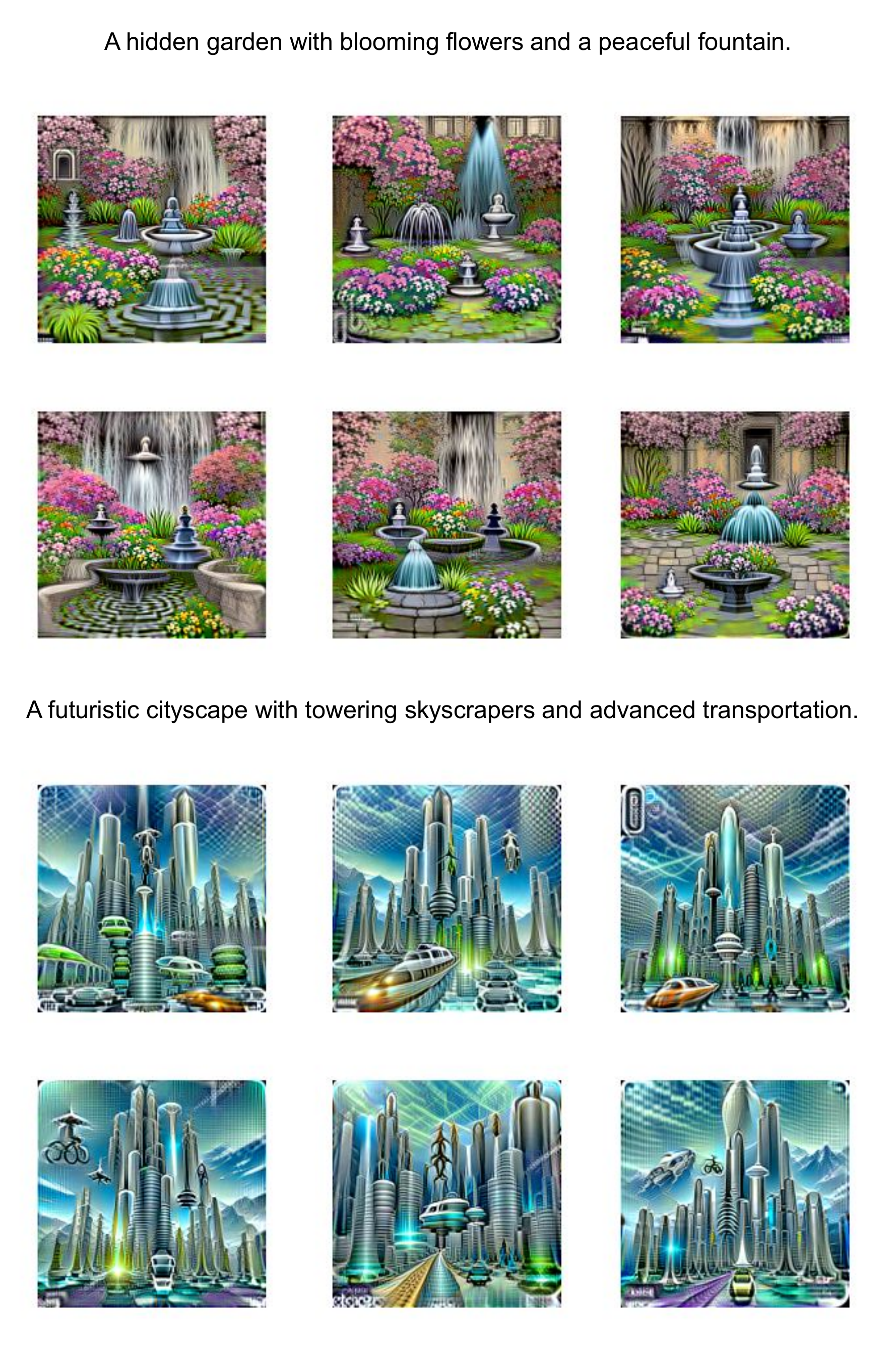}
    \caption{\textbf{CLIPAG generator-free text-to-image stochasticity}.
    We generate each textual description six times, to demonstrate the stochasticity introduces by our framework.}
    \label{fig:randomness}
\end{figure*}

Moreover, we aim to explore the level of stochasticity of our framework.
Our scheme includes two random steps that introduce randomness to the synthesis process -- the initialization and the random augmentations.
To better understand the variability that these mechanisms introduce, we generate the same caption several times and visualize such results in~\Cref{fig:randomness}.

\begin{figure*}[t]
    \centering
    \includegraphics[width=\linewidth]{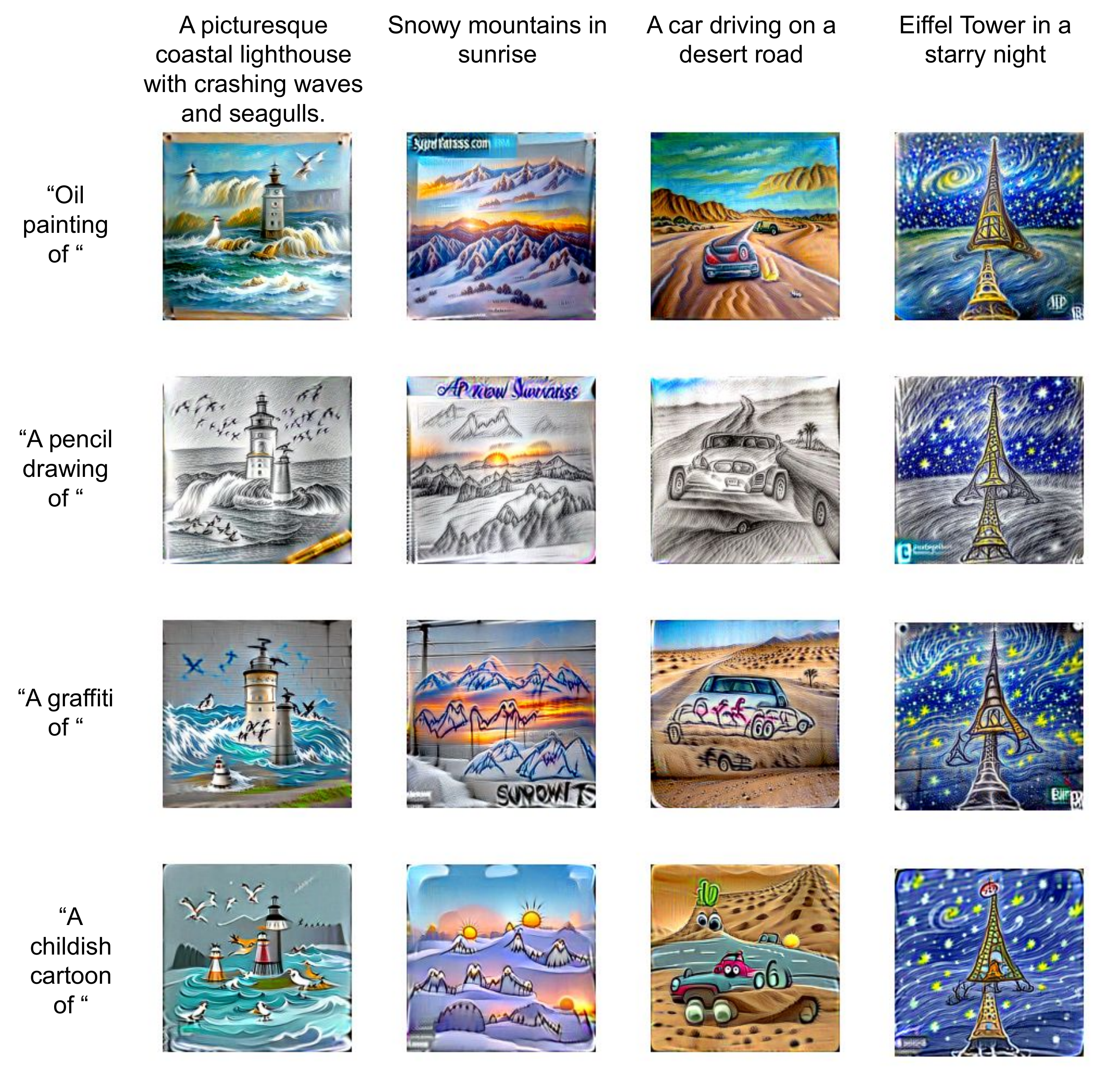}
    \caption{\textbf{CLIPAG generator-free text-to-image prefix effect}.}
    \label{fig:prompting}
\end{figure*}

Lastly, we investigate the effect of the chosen prompt on the generated image. Until now, we do not prepend to the target caption any guiding prefix. Now, we study the impact of adding such prefixes that describe the style of the desired image.
In particular, we consider the following prompts -- ``\texttt{oil painting of}'', ``\texttt{a pencil drawing of}'', ``\texttt{a graffiti of}'', and ``\texttt{a childish cartoon of}''.
We depict the results in~\Cref{fig:prompting}.
As can be seen, the prefix strongly determines the style of the generated images, strongly attesting to CLIPAG's capability in guiding towards different stlyes, although mainly trained on natural images.

\end{document}